\documentclass{article}

\PassOptionsToPackage{square, numbers, sort&compress}{natbib}

\usepackage[final]{neurips_2022}

\usepackage[utf8]{inputenc} 
\usepackage[T1]{fontenc}    
\usepackage{hyperref}       
\usepackage{url}            
\usepackage{booktabs}       
\usepackage{amsfonts}       
\usepackage{nicefrac}       
\usepackage{microtype}      
\usepackage{xcolor}         
\usepackage{graphicx}       
\usepackage{multirow}
\usepackage{amsmath}
\usepackage{enumerate}
\usepackage{listings}
\usepackage{caption}
\usepackage{subcaption}

\usepackage{array}
    \newcolumntype{P}[1]{>{\centering\arraybackslash}p{#1}}
    \newcolumntype{M}[1]{>{\centering\arraybackslash}m{#1}}

\newcommand{\purple}[1]{{\leavevmode\color{purple}#1}}

\let\OldTexttt\texttt
\renewcommand{\texttt}[1]{\purple{\OldTexttt{#1}}}

\newcommand{\dw}[1]{\multirow{2}{*}{#1}}

\title{Avalon: A Benchmark for RL Generalization Using Procedurally Generated Worlds}

\author{%
  Joshua Albrecht  \quad 
  Abraham J. Fetterman \quad
  Bryden Fogelman \quad 
  Ellie Kitanidis \quad \And
  Bartosz Wróblewski \quad
  Nicole Seo \quad
  Michael Rosenthal \quad 
  Maksis Knutins \quad \And
  Zachary Polizzi \quad
  James B. Simon \quad
  Kanjun Qiu \\ \\ \\
  Generally Intelligent\thanks{Correspondence to \OldTexttt{avalon@generallyintelligent.com}.} \\ \\
}

\begin{document}

\maketitle

\begin{abstract}
    Despite impressive successes, deep reinforcement learning (RL) systems still fall short of human performance on generalization to new tasks and environments that differ from their training. As a benchmark tailored for studying RL generalization, we introduce Avalon, a set of tasks in which embodied agents in highly diverse procedural 3D worlds must survive by navigating terrain, hunting or gathering food, and avoiding hazards. Avalon is unique among existing RL benchmarks in that the reward function, world dynamics, and action space are the same for every task, with tasks differentiated solely by altering the environment; its 20 tasks, ranging in complexity from \texttt{eat} and \texttt{throw} to \texttt{hunt} and \texttt{navigate}, each create worlds in which the agent must perform specific skills in order to survive. This setup enables investigations of generalization within tasks, between tasks, and to compositional tasks that require combining skills learned from previous tasks. Avalon includes a highly efficient simulator, a library of baselines, and a benchmark with scoring metrics evaluated against hundreds of hours of human performance, all of which are open-source and publicly available. We find that standard RL baselines make progress on most tasks but are still far from human performance, suggesting Avalon is challenging enough to advance the quest for generalizable RL.
\end{abstract}

\section{Introduction}\label{sec:introduction}
A central goal of reinforcement learning (RL) is to build systems that can master a spectrum of skills in environments as noisy and diverse as the real world.  While RL algorithms have matched or exceeded human performance on a number of narrowly-defined tasks such as Go and Atari \citep{Mnih++15, Go_Silver++17, Poker_BrownSandholm19, OpenAI_Five_19, Puigdom++20}, existing models typically fail to generalize to unseen tasks and environments, even when testing on environments drawn from the same distribution as training \citep{Zhang++18a, Zhang++18b, Farebrother++18}. The real-world setting is far more diverse and difficult than most existing RL benchmarks; agents must seamlessly interact in a highly variable 3D environment, with no access to state information or ability to rely on hard-coded discrete actions, requiring sensory inputs and a high-dimensional action space.

To aid the study of generalization in a more realistic setting that is still tractable for current RL algorithms, we present Avalon\footnote{Named after a mythical land of Arthurian legend whose etymology is ``the isle of fruit trees,'' since Avalon's worlds are mostly islands with fruit trees wherein the ultimate goal is to find and eat the fruit.}: a benchmark and high-performance simulator. Avalon is a 3D open-world survival game in which agents must navigate terrain, find food, use tools, hunt prey, and avoid predators and other dangers. Highly diverse 3D worlds are procedurally generated with many biomes, plants, items with distinct properties, and animals with unique behaviors. In any particular world, the agent’s sole objective is to survive for as long as possible by finding and eating all food while avoiding hazards. Avalon agents are embodied and receive visual and proprioceptive input, forcing them to learn representations rather than relying on latent state information that would be inaccessible in the real world.

\begin{figure}
  \centering
    \includegraphics[width=1.0\linewidth]{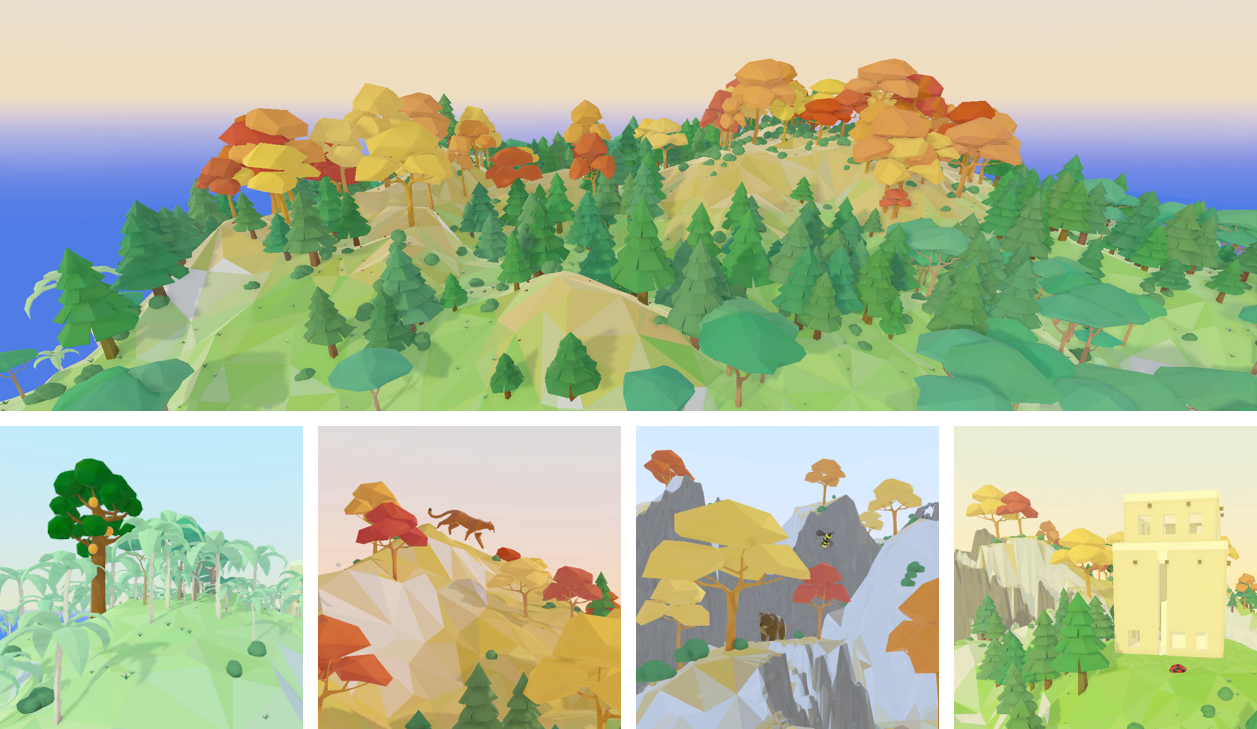}
    \caption{\textbf{
    Avalon is an open-world survival game requiring an agent to navigate hazards and find food.}
    Top: aerial view of a procedurally-generated world.
    Bottom row, left to right: examples of fruit to consume, predators to avoid, terrain to navigate, and buildings to explore.
    }
    \label{fig:figure_label}
\end{figure}

We formulate Avalon as a multi-task RL benchmark in which each task is shaped by environmental pressures rather than a task-dependent reward function. This enables all tasks to have shared dynamics and reward, making generalization more feasible compared to disjoint task spaces such as Atari's \cite{ALE_12} where reduced or even negative transfer learning has been noted \cite{Parisotto++15} \cite{Rusu++15}. Twenty tasks ranging in complexity from \texttt{eat} and \texttt{throw} to \texttt{hunt} and \texttt{survive} are included in the benchmark. A \emph{world generator} for each task carefully alters a world so that certain skills are required to complete the task. For example, the \texttt{climb} generator only spawns food in places that cannot be accessed without climbing. Thus, Avalon’s tasks are effectively a set of functions for procedurally generating environments that force the agent to learn the requisite skill to survive. 

Avalon facilitates the exploration of a variety of forms of generalization:

\textbf{To unseen tasks that are structurally similar to previous tasks}. All tasks in Avalon share the same transition dynamics, action space, observation space, and simple reward based on the agent's energy level. Environment variation between tasks and within tasks is executed through the same mechanism i.e. sampling from a subset of the full distribution of possible worlds. Indeed, one could imagine widening the scope of any task-specific world generator until it encompasses some or all other tasks. This notion of task ``adjacency'' enables the agent to exploit shared structure between tasks.

\textbf{To unseen combinations of tasks}. In addition to 16 basic generators that map to basic skills, four compositional generators create worlds that require the agent to exercise random combinations, permutations, and variations of those skills. These compositional generators can be used to evaluate the agent's performance on unseen tasks that require many skills at once and provide a new setting to study the generalizability and compositionality of learned skills. In other multi-task setups, compositional tasks are often difficult to express. For example, multi-term reward functions quickly become cumbersome, and designing goal states that require complex sequential tasks is challenging.

\textbf{To unseen environments within tasks}. Unlike many popular RL benchmarks (e.g. Atari and DeepMind Control \cite{DMLab_Beattie++16}) where the test environment is identical to the train environment, Avalon has significantly more variation between worlds sampled from the same distribution. Even compared to other procedurally generated benchmarks like ProcGen \cite{ProcGen_19} and MineRL \cite{MineRL_19}, Avalon contains dramatically more factors of variation, each of which can be individually and finely controlled. Agents in Avalon must generalize to IID sampled test environments in a setting with significantly more realism and complexity than comparable benchmarks.

Avalon’s continuous action space maps to a virtual reality (VR) headset and controllers. Our choice of observation and action space allows us to measure human performance by recording demonstrations using a VR headset. We include this dataset of human VR actions, with 215 hours of human playthroughs on 1000 worlds, which may be of independent research interest. 

Our benchmark includes a number of state-of-the-art RL algorithms as baselines. In Section~\ref{sec:experiments}, we show their performance on our tasks and highlight training mechanisms enabled by our world generation setup that help them achieve non-trivial (though still far short of human) performance.

As a final contribution, we are fully open-sourcing our environment, which represents a significant engineering effort and includes the following key contributions:

\begin{itemize}

\item \textbf{Efficiency}. At 7000 SPS (steps per second), Avalon is on-par with the fastest comparable simulator \cite{Habitat_19}.

\item \textbf{Usability}. Avalon is fully open-source, built on Godot (a free game engine with an intuitive visual editor), and includes training scripts and debugging tools.

\item \textbf{Configurability}. Control over the many factors of variation in the world generators enables users to fully control the training procedure and learning curriculum.

\end{itemize}

Our hope is that the Avalon benchmark and simulator will serve as useful tools for the RL community in the quest to build more general, capable learning systems.

\section{Related work}\label{sec:related-work}
Avalon is the only benchmark where 1) agents learn from high-dimensional inputs in 3D procedurally generated worlds with a continuous action space, 2) the observation space, action space, transition dynamics, and reward are held constant across all tasks, 3) a very large number of factors of variation can be finely controlled in order to isolate and explore specific types of generalization, and 4) the underlying simulator is very fast and easy to use. To achieve these goals, Avalon builds on many ideas from previous works, discussed below.

Games have historically been the gold standard for benchmarking reinforcement learning methods \citep{Go_Silver++17, Poker_BrownSandholm19, OpenAI_Five_19}. The Arcade Learning Environment \cite{ALE_12} and DeepMind Lab \cite{DMLab_Beattie++16} are early examples of multi-task benchmarks that encourage mastery across several tasks. On these benchmarks, overfitting due to trajectory memorization \cite{Justesen++18} and poor or negative transfer across tasks \citep{Parisotto++15, Rusu++15} remain issues.

Procedurally generated benchmarks such as ProcGen \cite{ProcGen_19}, MineRL \cite{MineRL_19}, Malmo \cite{Malmo_16}, and MetaDrive \cite{MetaDrive21} aim to address per-task overfitting by varying environmental factors such as the background and placement of objects and obstacles. However, most rely on a single random seed for procedural content generation or allow users to vary only a few (usually discrete) parameters. MineRL and Malmo also run at less than 50 steps per second on a single GPU (over 100 times slower than Avalon), which makes their use for large experiments slow and expensive.

Other works such as Meta-World \cite{MetaWorld_19} attempt to address poor task transfer by unifying the observation space, action space, and transition dynamics across all tasks and adding controllable parametric variation of object and goal positions in order to increase shared structure across tasks. However, these environments rely on direct access to state information and include per-task handcrafted dense rewards, both of which limit applicability to the broad set of tasks that general agents are expected to solve.

Other benchmarks such as Crafter \cite{Crafter_21}, Obstacle Tower \cite{ObstacleTower_19}, and the Animal-AI Environment \cite{AnimalAI_19} avoid per-task rewards by defining a single reward function. These environments all have discrete action spaces and are limited in terms of diversity. Crafter is a 2D world; Obstacle Tower has few game mechanics: opening doors, picking up keys, pushing blocks; and Animal-AI tasks exist in a flat arena with a dozen object options.

Several other environments have substantially more diversity. MiniHack \cite{MiniHack_21} is a sandbox for designing environments in the text-based game NetHack \cite{NetHack_20}, which has a randomized system of dungeons with many creatures and items. However, mastery of NetHack is elusive even for humans and requires extensive game-specific knowledge, making it ill-suited to generalization research, and NetHack’s symbolic inputs do not test an agent’s ability to learn visual representations. By contrast, XLand \cite{XLand_21} is a 3D multi-agent environment with visual inputs and a single reward function across tasks. Agents in XLand operate in a low-dimensional, discrete action space and are provided with explicit state information about the goals and game predicates, which severely limits the tasks that can be expressed in this constrained symbolic system. Additionally, XLand is not publicly released, so it cannot be used as a benchmark.

A separate class of related work includes photorealistic embodied AI environments such as Habitat Lab \cite{Habitat_19}, GibsonEnv \cite{Gibson_18}, Megaverse \cite{Megaverse_21}, RFUniverse \cite{RFUniverse_22} and BEHAVIOR \cite{BEHAVIOR_21}, 3D simulators such as ThreeDWorld \cite{ThreeDWorld_20}, and dataset generators such as Kubric \cite{Kubric_22}. These are valuable contributions for embodied AI and visual representation learning, but they are not designed as benchmarks for exploring RL generalization. Additionally, most of these simulators run at less than 100 steps per second (SPS) on a single GPU, which is prohibitively slow for RL research. Habitat Lab runs significantly faster (around 8,000 SPS on a single GPU), but has a limited action space, limited allowed physical interactions, and no procedural generation, making it ill-suited as a benchmark for generalization.

\section{RL interface}\label{sec:interface}
The Avalon RL environment conforms to the standard OpenAI Gym environment interface \citep{gym} and is built on top of the open-source Godot game engine \citep{godot} and Bullet physics engine \citep{bullet}. Below, we present some details of this environment and the design decisions that led to it.

\subsection{Game mechanics}

\begin{figure}
  \centering
    \includegraphics[width=1.0\linewidth]{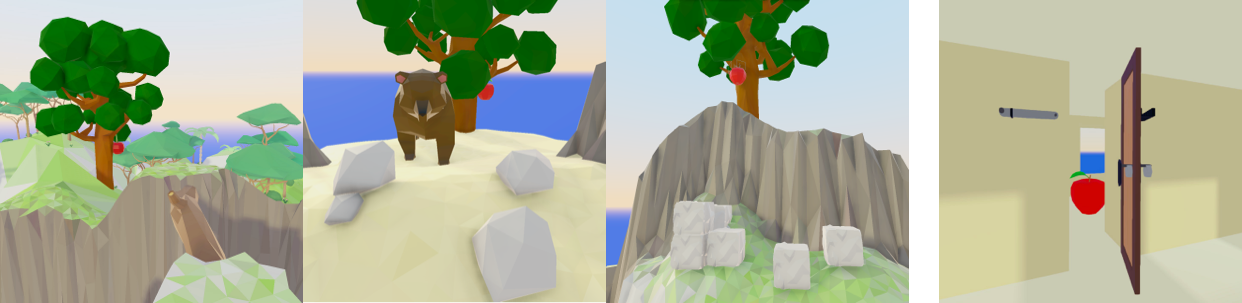}
    \caption{Example tasks \texttt{bridge}, \texttt{fight}, and \texttt{stack} (left). World generators for each task shape the environment to require completion of the task to reach food. Tasks can be outdoor or inside buildings in the environment. Buildings also enable \texttt{open} (right), where doors with complex locking mechanisms must be solved to get to food.}
    \label{fig:task_examples}
\end{figure}

Avalon is a 3D open-world survival game in which players must overcome obstacles and hazards while acquiring food. Gameplay consists of a series of episodes, where each episode corresponds to a unique world, which is generated by a task-specific world generator.

A key aspect of Avalon is that special subsets of the environment distribution map to distinct tasks. Environmental pressures arising from the presence and placement of various interactable objects, enemies, and even terrain features demand the execution of a broad array of navigation and object interaction skills. For example, the existence of a deep chasm between the player and the food forces the player to create a bridge, or food may be placed in a region that is inaccessible without climbing. Figure~\ref{fig:task_examples} shows some examples of generated worlds for a variety of tasks.

Another critical part of Avalon is the diversity of gameplay mechanics and worlds that can be generated. There are 14 biomes, 13 plants, 20 interactable items, and 17 animals. Items include food, weapons, elements such as boxes for stacking, doors, and so on. Animals include both predators and prey, where prey are edible and can be consumed by the agent as food. All animals and food have completely distinct behaviors (for example, jaguars aggressively pursue the player and can climb trees, coconuts must experience a certain force to be opened, etc). See Appendix~\ref{app:biomes_objects_etc} for a complete description of each entity included in the game.

While the setting of Avalon is inspired by the environment in which humans evolved, it includes a number of simplifications designed to accommodate current RL systems. For example, most objects are significantly larger than their real-life counterparts because current RL algorithms are prohibitively expensive to train with very high-resolution images. As another mechanism to allow for low-resolution agents, we guarantee that food is always found near buildings or certain large trees, both of which are visible from far away. 

\subsection{Rewards}

Regardless of the task, the goal of players in Avalon is to survive for as long as possible. Both human players and RL agents have a single scalar “energy” value, and when this reaches zero, the episode ends. The only way to gain energy is to eat, and thus acquiring food is the implicit goal of every level. Energy can be lost from attacks by predators or falling too far.

In order to represent these mechanics to the agent, the reward from the environment is simply calculated as the change in energy at each time step. While the reward is technically dense, in that the agent observes small changes in energy at each frame, it is effectively a sparse reward, since positive reward is attained only when the agent eats food i.e. has successfully performed the task.

To encourage efficiency and prevent erratic motion, small energy costs can be associated with movement for agents during training.
More details on the exact calculations for the reward function and time limits in each episode can be found in Appendix~\ref{app:reward}.

Though this (effectively) sparse reward setting is challenging, we note that dense rewards usually need to be tailored to each task and often rely on hidden state information. 
However, it is straightforward for a user to adapt Avalon to employ a dense reward if desired.

\subsection{Observations}

While the state data for our simulation is easily accessible, we refrain from providing any ground truth information in the observations that is not accessible to human players.
The agent receives egocentric visual input in the form of a 96 $\times$ 96 $\times$ 4 RGBD tensor. Just like human players, the agent is embodied and can see its own hands in its visual display.
The agent is also provided with basic proprioceptive awareness: its input includes the framewise change in position and orientation for its body, the position and orientation of each of its hands (relative to its body), boolean indicators for whether each hand is within grasping range of (or currently grasping) an object, the current value of its energy, and the number of frames remaining until the episode times out. See Appendix~\ref{app:observation} for the exact list of observed variables.

\subsection{Actions}

The embodied agent can move through and interact with its environment using its head, body, and hands. The primary action space is 21-dimensional and roughly maps to a virtual reality headset and controllers. It is made of $3 \times (3 + 3)$ translational plus rotational degrees of freedom, as well as $2 \times 1$ binary grasp actions and 1 discrete jump action. 
We also provide a reduced 9-dimensional action space which corresponds to mouse and keyboard controls. Almost all tasks can be accomplished in this reduced action space.
See Appendix~\ref{app:action} for the exact action space definition.

\subsection{Simulator}

Unlike many popular RL environments, Avalon is not built on top of an existing game but rather constructed from the ground up to optimize for speed, accessibility, and, above all, scientific value. This has enabled several benefits that would otherwise be impossible:

\begin{itemize}

\item Avalon simulates roughly 7,000 SPS, orders of magnitude faster than most other simulators, and similar to Habitat-Sim, the fastest comparable simulator. The simulator runs without a display, making it even more efficient and easy to work with.

\item Avalon was designed for the purposes of ML research, and thus benefits from a number of debugging capabilities. Firstly, it is deterministic; playing back the same set of actions from the same starting seed results in identical output. Secondly, the ground truth states of all objects can be logged continually, making it highly inspectable.

\item Avalon is made using the Godot game engine \citep{godot}, which is fully open-source and cross-platform (it runs on Windows, Linux, Mac, and several popular VR headsets). Godot is simple and lightweight, with a modest $\sim$30MB install size, and has a clean asset pipeline that supports most common formats. It boasts a full-featured visual editor, debugger and profiler that make it very easy to work with. All of our code is written in either Python or the Python-esque Godot scripting language, so it is straightforward for researchers to modify and extend, especially given the vibrant community of professional and hobbyist developers that also use Godot.

\end{itemize}

Godot is released under an MIT license, while the rest of our code is released under the GPL license.

\section{Procedural environment generation}\label{sec:proc_gen_env}
Avalon contains a sophisticated system for generating worlds that gives researchers fine-grained control over every aspect of variation—such as ruggedness of terrain, height of cliffs, density of predators, prey, and plants, etc.—while enabling extremely high diversity. This system empowers researchers to isolate and explore many specific aspects of generalization, ranging from the most straightforward IID setting to OOD settings like the creation of complex compositional tasks or even entirely new tasks. It also enables a simple difficulty-based curriculum that accelerates learning for our included baseline systems (described in more detail in Section~\ref{sec:experiments}).

Avalon supports easily scaling the diversity of the training distribution, allowing for exploration of basic IID generalization, where the test world distribution matches the training distribution. In our benchmark setting, we primarily vary terrain height and geography while leaving most diversity turned off (e.g. we don’t vary colors or models of trees, landscape, food, prey, or predators, nor do we vary objects in size, shape, or weight, etc.). As RL systems become more capable, this diversity can be dramatically increased to provide more challenging IID generalization settings.

To enable investigation of generalization between similar tasks, Avalon’s task-specific world generators carefully alter each procedurally generated world to require the use of particular skills. For example, the generator can raise a section of terrain to surround food with a cliff or insert an unclimbable chasm to require jumping. Our hope is that this setup enables better transfer between similar tasks, as many tasks allow for multiple solutions and the use of multiple skills, especially at lower difficulties.

In order to address a more compositional form of OOD generalization, we have designed our task-specific world generators so that they can be applied to a single world to create a sequence of tasks that must be performed. For example, our “jump” generator creates a natural-looking chasm that can only be jumped across at a certain point, while our “climb” generator creates a ring of sloped terrain that can only be climbed along a certain path. By simply composing one of these obstacle “rings” inside the other, we can create a world in which the player must accomplish both.

Our fine-grained control over diversity in world generation can also be used to directly ask about OOD generalization under small shifts in the distribution of environments. All variation in Avalon’s world generation is finely controllable: the sizes and colors of all objects, the number of predators, prey, food, etc in a task, the distance of a gap that must be jumped across, etc (see Appendix~\ref{app:FOV} for a full list). This allows for simple, continuous ablations as any individual factor is varied from the training setting, all without any need to create or import any manual art assets.

\section{Benchmark}\label{sec:benchmark}
\subsection{Tasks}

Avalon consists of 20 distinct tasks testing a range of navigation and object manipulation skills in diverse environments.
This list includes 16 ``basic'' tasks (\texttt{eat}, \texttt{move}, \texttt{jump}, \texttt{climb}, \texttt{scramble}, \texttt{descend}, \texttt{throw}, \texttt{hunt}, \texttt{fight}, \texttt{avoid}, \texttt{push}, \texttt{stack}, \texttt{bridge}, \texttt{open}, \texttt{carry}, and \texttt{explore}) and four ``compositional'' tasks (\texttt{navigate}, \texttt{find}, \texttt{gather}, and \texttt{survive}).
Each task has the same agent goal --- acquiring food while avoiding hazards --- but the worlds generated for each task are structured such that attaining this goal requires the skill being tested.

Worlds for the 16 basic tasks include a single food item that must be reached by overcoming at most a single obstacle.
The \texttt{eat} task is the easiest: the food is created near the agent, which needs to merely grab it and bring it to its head.
The other basic tasks build upon this in various ways. For example, \texttt{avoid} requires evading a predator to reach the food, \texttt{hunt} requires hunting moving prey, and \texttt{stack} requires stacking objects to reach food on a high ledge.

Worlds for the four compositional tasks are designed to require the sequential use of several basic skills:
\texttt{navigate} places a series of basic obstacles between the player and the food,
\texttt{find} is the same but the food is not guaranteed to be visible from the agent's starting location,
\texttt{gather} has multiple fruits to find, and
\texttt{survive} includes both fruit and prey animals.
\texttt{survive} is a sort of ``final exam'' focused on breadth whose worlds can contain the elements of every other task.

See Appendix~\ref{app:tasks} for a complete definition of every task.

\subsection{Training and evaluation protocols}
\label{sec:training-protocols}

We outline four settings in which agents can be trained and evaluated within Avalon:

\begin{itemize}
\item \textbf{Multi-Task, Train Basic (MT-TB)}: Train on the 16 basic tasks and evaluate on all 20 tasks. In this version of the multi-task setup, agents have seen each of the intermediary tasks but have never seen them composed together until test time.
\item \textbf{Multi-Task, Train All (MT-TA)}: Train and evaluate on all 20 tasks.
\item \textbf{Multi-Task, Train Compositional (MT-TC)}: Train on the four compositional tasks and evaluate on the full set of 20 tasks.
\item \textbf{Single Task, Basic (ST-B)}: Train and evaluate on each of the 16 basic tasks separately.
\end{itemize}

Agents are trained with an adaptive curriculum (Section~\ref{subsec:curriculum}) such that the number of worlds seen in a given amount of training time is variable.  While we present a particular curriculum-based training procedure, users of the benchmark are encouraged to explore alternative training procedures. 

Agents are evaluated on a fixed set of 50 worlds for each relevant task, as indicated above. For example, the MT-CG and MT-S settings evaluate on all 20 tasks, and thus use the full set of 1,000 possible evaluation worlds. Tasks were randomly generated and not manually curated, except to replace levels that were reported as practically impossible.  Upon investigation only one level was actually impossible,  representing 0.1\% of the originally generated levels. We also ensured that the evaluation worlds fully represented their underlying task generators. For example, the set of evaluation tasks for ``eat'' includes at least one of every possible type of food. See Appendix~\ref{app:eval_worlds} for more details about the exact distribution of evaluation worlds, including the replaced worlds.

\subsection{Human performance}

To understand the difficulty of each of our tasks, we collected human player data for all evaluation levels. Approximately 215 hours of VR gameplay were recorded from 32 participants drawn from a pool of volunteers whose familiarity ranged from zero to significant experience playing similar games. For each of the $50 \times 20 = 1000$ evaluation levels, scores were averaged across at least 5 different players. Humans were provided with two practice levels for each task (not in the evaluation set), as well as basic instructions about the game mechanics (see Appendix~\ref{app:humans} for details on the information provided and human data collection procedure).

Ground truth control data was recorded for each player on each level, and is available on our main website under a CC BY-SA license. We do not recommend behavior cloning from this data and testing on the fixed evaluation set; however, researchers can easily generate new evaluation levels from the same distribution for testing any such networks. The human data may also be interesting to inspect for trends, training effects, or unrelated research into VR human gameplay.

\subsection{Performance metrics}

Given a set of human player data, raw rewards can be converted into scores $S$ by normalizing such that the average human performance is $1.0$ and the performance of a random agent is $0.0$. We recommend two different aggregations of these scores. The first and more traditional aggregation is to simply take the average of these scores. The second and more robust aggregation is the optimality gap \cite{Agarwal++21}. By plotting the cumulative frequency of these scores $P(S>x)$ across all runs, the optimality gap expresses how consistently the agent achieves performance above some fraction of human performance. See Appendix~\ref{app:scoring} for details on how scores are calculated.

\section{Experiments}\label{sec:experiments}
\subsection{Difficulty curriculum}
\label{subsec:curriculum}

Due to the variety and scope of tasks and environments in our benchmark, we found that agents failed to learn when training on worlds sampled uniformly at random. Because the agent is effectively only given sparse positive rewards (i.e. if it finds and eats food), it is difficult for agents to make progress on more difficult tasks before improving at the \texttt{eat} and \texttt{move} tasks. For most tasks, uniform random sampling of worlds leads to a very low probability of creating worlds where random exploration policies get any reward. 

As one approach to overcoming these issues, we use a simple ``difficulty''-based curriculum. We employ a simple form of Automated Curriculum Learning \cite{Portelas++20} similar to the approaches in \citep{Kanitscheider++21, OpenAI_RobotHand_19}. When the agent succeeds (fails) at a task, the maximum difficulty of future generated worlds for that task is increased (decreased) slightly. See Appendix~\ref{app:curriculum} for more details.

\subsection{Baselines}

We trained IMPALA \citep{espeholt:2018-impala, kuttler:2019-torchbeast}, PPO \citep{schulman:2017-ppo, raffin:2021-sb3}, and DreamerV2 \citep{dreamerV1, dreamerV2} on our benchmark using the training protocols outlined above, including the difficulty curriculum. We chose IMPALA and PPO as baselines because they are among the simplest SOTA algorithms on which many other SOTA algorithms are based, and tend to be robust across problems. We chose DreamerV2 as a representative model-based baseline. DreamerV2 is used without discrete latents and without mixed precision training. No worlds are seen twice during training. Hyper-parameters were tuned via a combination of Bayesian optimization \citep{jones:1998-efficient-opt} and Natural Evolution Strategies (NES) \citep{wierstra:2014-nes} using runs with a smaller number of steps, and are included in Appendix~\ref{app:hyperparams}. Scores are the average of 5 runs (for 50m step results) or 3 runs (for the 500m step results). See Appendix~\ref{app:compute} for details about the machines used for training.

\subsection{Results and discussion}
\begin{table}
  \caption{Average agent scores on all Avalon tasks. Rows show scores on basic tasks (first section), compositional tasks (second section), and aggregations of tasks (third section), normalized such that mean human performance is one. The header indicates the algorithm, the total number of environment steps and whether the training curriculum is used.}
  \label{tab:scores}
  \centering
    \begin{tabular}{llllll}
    \toprule

    \textbf{Task} & \textbf{PPO} & \textbf{Dreamer} & \multicolumn{3}{c}{\textbf{IMPALA}} \\
    
    \cmidrule{4-6}
        &   50m steps  &   50m steps &  \multicolumn{1}{c}{50m steps}  & \multicolumn{1}{c}{500m steps}   & \multicolumn{1}{c}{50m steps}\\
        &   With curr. &   With curr.  &  \multicolumn{1}{c}{With curr.}  & \multicolumn{1}{c}{With curr.}   & \multicolumn{1}{c}{No curr.}   \\
\midrule
 \texttt{eat}       & 0.708 $\pm$ 0.067 & 0.664 $\pm$ 0.065  & 0.716 $\pm$ 0.062 & 0.731 $\pm$ 0.097        & 0.001 $\pm$ 0.001                 \\
 \texttt{move}      & 0.311 $\pm$ 0.062 & 0.364 $\pm$ 0.071  & 0.399 $\pm$ 0.062 & 0.409 $\pm$ 0.076        & 0.000 $\pm$ 0.000                 \\
 \texttt{jump}      & 0.220 $\pm$ 0.050 & 0.234 $\pm$ 0.058  & 0.309 $\pm$ 0.056 & 0.287 $\pm$ 0.072        & 0.000 $\pm$ 0.000                 \\
 \texttt{climb}     & 0.193 $\pm$ 0.043 & 0.227 $\pm$ 0.051  & 0.229 $\pm$ 0.049 & 0.332 $\pm$ 0.074        & 0.000 $\pm$ 0.000                 \\
 \texttt{descend}   & 0.179 $\pm$ 0.043 & 0.290 $\pm$ 0.058  & 0.173 $\pm$ 0.044 & 0.222 $\pm$ 0.059        & 0.000 $\pm$ 0.000                 \\
 \texttt{scramble}  & 0.306 $\pm$ 0.054 & 0.422 $\pm$ 0.058  & 0.467 $\pm$ 0.062 & 0.576 $\pm$ 0.070        & 0.000 $\pm$ 0.000                 \\
 \texttt{stack}     & 0.091 $\pm$ 0.036 & 0.126 $\pm$ 0.043  & 0.130 $\pm$ 0.036 & 0.121 $\pm$ 0.055        & 0.000 $\pm$ 0.000                 \\
 \texttt{bridge}    & 0.049 $\pm$ 0.027 & 0.121 $\pm$ 0.045  & 0.076 $\pm$ 0.029 & 0.095 $\pm$ 0.049        & 0.000 $\pm$ 0.000                 \\
 \texttt{push}      & 0.113 $\pm$ 0.039 & 0.160 $\pm$ 0.053  & 0.150 $\pm$ 0.043 & 0.128 $\pm$ 0.054        & 0.000 $\pm$ 0.000                 \\
 \texttt{throw}     & 0.000 $\pm$ 0.000 & 0.000 $\pm$ 0.000  & 0.000 $\pm$ 0.000 & 0.000 $\pm$ 0.000        & 0.000 $\pm$ 0.000                 \\
 \texttt{hunt}      & 0.043 $\pm$ 0.024 & 0.063 $\pm$ 0.028  & 0.071 $\pm$ 0.029 & 0.129 $\pm$ 0.051        & 0.000 $\pm$ 0.000                 \\
 \texttt{fight}     & 0.199 $\pm$ 0.052 & 0.336 $\pm$ 0.076  & 0.235 $\pm$ 0.052 & 0.303 $\pm$ 0.075        & 0.000 $\pm$ 0.000                 \\
 \texttt{avoid}     & 0.493 $\pm$ 0.170 & 0.515 $\pm$ 0.118  & 0.603 $\pm$ 0.159 & 0.582 $\pm$ 0.102        & 0.000 $\pm$ 0.000                 \\
 \texttt{explore}   & 0.193 $\pm$ 0.047 & 0.190 $\pm$ 0.048  & 0.213 $\pm$ 0.048 & 0.252 $\pm$ 0.069        & 0.000 $\pm$ 0.000                 \\
 \texttt{open}      & 0.055 $\pm$ 0.024 & 0.126 $\pm$ 0.041  & 0.097 $\pm$ 0.034 & 0.101 $\pm$ 0.046        & 0.000 $\pm$ 0.000                 \\
 \texttt{carry}     & 0.073 $\pm$ 0.031 & 0.066 $\pm$ 0.028  & 0.089 $\pm$ 0.032 & 0.122 $\pm$ 0.057        & 0.000 $\pm$ 0.000                 \\
\midrule
 \texttt{navigate}  & 0.000 $\pm$ 0.000 & 0.000 $\pm$ 0.000  & 0.012 $\pm$ 0.010 & 0.040 $\pm$ 0.032        & 0.000 $\pm$ 0.000                 \\
 \texttt{find}      & 0.002 $\pm$ 0.003 & 0.000 $\pm$ 0.000  & 0.015 $\pm$ 0.014 & 0.013 $\pm$ 0.017        & 0.000 $\pm$ 0.000                 \\
 \texttt{survive}   & 0.043 $\pm$ 0.013 & 0.044 $\pm$ 0.014  & 0.050 $\pm$ 0.015 & 0.085 $\pm$ 0.028        & 0.000 $\pm$ 0.000                 \\
 \texttt{gather}    & 0.021 $\pm$ 0.010 & 0.021 $\pm$ 0.012  & 0.030 $\pm$ 0.010 & 0.032 $\pm$ 0.014        & 0.000 $\pm$ 0.000                 \\
\midrule
 all basic & 0.202 $\pm$ 0.017 & 0.244 $\pm$ 0.016  & 0.247 $\pm$ 0.016 & 0.274 $\pm$ 0.019        & 0.000 $\pm$ 0.000                 \\
 all comp. & 0.017 $\pm$ 0.004 & 0.016 $\pm$ 0.005  & 0.027 $\pm$ 0.007 & 0.042 $\pm$ 0.013        & 0.000 $\pm$ 0.000                 \\
 all       & 0.165 $\pm$ 0.014 & 0.199 $\pm$ 0.012  & 0.203 $\pm$ 0.013 & 0.228 $\pm$ 0.015        & 0.000 $\pm$ 0.000                 \\

\bottomrule
\end{tabular}
\end{table}

We show average scores for the MT-TB setting in Table~\ref{tab:scores}. While all algorithms are able to achieve non-zero performance on most tasks, they fall far short of human performance, and show particularly low performance when testing on the four compositional tasks. It should be noted that non-zero performance is sometimes indicative of unexpected strategies; for example, we observe agents learning to open doors by jumping while grabbing the latch rather than moving their hand to lift the latch, thus reliably opening doors without necessarily understanding doors. 

Table~\ref{tab:scores} also includes the results of longer training and ablating the curriculum learning component. The longer training shows results comparable to the 50m step training, indicating convergence of IMPALA, although learning curves in Appendix~\ref{app:other_results} indicate some tasks might still see improvement with longer training. Using no curriculum results in scores that are no better than random. 

Due to space constraints, the per-task optimality gap scores are reported in Appendix~\ref{app:other_results} (see Table~\ref{tab:optimality_gap_scores}) along with the scores from the other train-evaluate settings (MT-TA, MT-TC, ST-B) and additional results. Comparisons of the different settings (see Table~\ref{tab:taskwise_agent_scores} and Table~\ref{tab:taskwise_optimality_gap_scores}) provide multiple lenses through which to view generalization. For example, we find that several tasks such as \texttt{climb}, \texttt{descend}, and \texttt{avoid} are too challenging for the agent to learn in the single-task setting (ST-B), yet they can be partially mastered in the multi-task settings (MT-TB, MT-TA) after pre-training on other tasks, suggesting the shared structure of the tasks provides transfer learning benefit. Furthermore, we find that agents trained on only the 16 basic tasks (MT-TB) perform as well as or better than agents trained on all 20 tasks (MT-TA), even on the compositional tasks themselves which the agent has never seen in training under MT-TB.

\section{Limitations}\label{sec:limitations}

While Avalon as it is today enables targeted exploration of many types of generalization in RL, there are a number of limitations. First, given the scope of the diversity possible in our environments, there are sure to be bugs and unintended results (for example, some generated worlds are impossible to complete). Another limitation is fundamental to procedural generation; while the generated worlds are quite complex, they are still far less complex than the real world, and as such, success on Avalon should not be taken as evidence that generalization is “solved” in the general case. Finally, due to limited time, we were only able to run a small number of baseline algorithms. We intend to publish updated baselines and results as they become available.

\section{Future work}\label{sec:future_work}

As it exists today, Avalon enables a variety of experiments, from the generalization settings highlighted here to better curriculum designs and unsupervised skill discovery. Avalon is also highly extensible by design. Users can easily create entirely new game mechanics, enabling RL research beyond the simple navigation and object manipulation tasks we have presented here.

We plan to extend Avalon in a number of ways, including by creating more tasks and more variety within each task. We also plan to include more sophisticated environment interactions to encourage tool making and tool use, as well as longer time-horizon mechanics such as rest, day-night cycles, and regrowing food to encourage the creation of RL agents that can deal with longer time horizons. Finally, we also hope to eventually include multi-agent components as well.

\section{Conclusions}\label{sec:conclusions}

We created Avalon to address the need for a better benchmark for RL generalization and robustness. By creating a diverse set of tasks solely via environmental variation, Avalon is able to create challenging worlds that share task structure and world dynamics, hopefully encouraging the creation of more powerful and generally capable RL agents. The tasks in the Avalon benchmark are quite challenging for existing systems, despite being relatively trivial for people. Despite this, our training procedures provide a starting point with reasonable performance, on which others can improve. We hope our simulator and benchmark will serve as a foundational piece of infrastructure for future research on generalization, exploration, and other topics that are under-served by existing benchmarks.

\begin{ack}
    The authors would like to thank the following individuals for invaluable discussions and feedback on this benchmark: Anwar Bey, Tom Brown, Michael Chang, Shreyas Kapur, Andrew Lampinen, Joel Lehman, Rosanne Liu, Luke Melas-Kyriazi, John Schulman, Elias Wang, Yi-Fu Wu, and Wojciech Zaremba.
    
    The authors are also grateful for the hard work and adventurous spirit of our human Avalon players: Meera Balakumar, Akshiv Bansal, Michael Bonanni, Andrew Cote, Rob Courtice, Dane Cross, Todd Dabney, Samkeliso Dlamini, Alejandra Encinas, Lincoln Scott Fuller, Amanda Gabbara, Nick Gabbara, Maddy Gaffaney, Gjon Gjeloshaj, Perry Goldstein, Cecilia Goss, Patrick Hoon, Schuyler Howe, Asmeret Jafarzade, Keil Miller Joseph, Luke Juusola, John Lindstedt K., Yad Konrad, Cory Li, Terrence Lucero, Kathryn Martin, Annie Melton, Brian Demeyer Michael, Kevin Multani, Jeremy Nelson, Carol Ng, Sam O’Donnell, Sam Parks, Divyesh Patel, Matthias Pauthner, Dominick Pierre-Jacques, Maria Polizzi, Nathan Ravenel, Roy Rinberg, Snigdha Roy, Katie Sapko, Phillip Seo, Brian Smiley, Derek Tam, Kaspars Vandans, Helen Wei, Christina Zhu.

\end{ack}

\bibliographystyle{abbrvnat}
\bibliography{references}

\clearpage

\appendix

\section*{Supplementary Materials: Content Overview}

\begin{itemize}
    \item Appendix \ref{app:biomes_objects_etc} contains details about each game element.
    \item Appendix \ref{app:reward} contains details about the reward function and starting energy levels.
    \item Appendix \ref{app:observation} contains a more detailed specification of the observation space.
    \item Appendix \ref{app:action} contains a more detailed specification of the action space.
    \item Appendix \ref{app:FOV} contains an overview of the different types of factors of variation in the generated worlds, with pointers to the full documentation in the code.
    \item Appendix \ref{app:tasks} contains a brief description of the mechanics of each of the 16 basic tasks and four compositional tasks.
    \item Appendix \ref{app:eval_worlds} contains more details about the exact procedure for generating and selecting the worlds used for evaluation.
    \item Appendix \ref{app:humans} contains a description of the information that was provided to the human participants as well as other details about the human data collection.
    \item Appendix \ref{app:curriculum} contains more details about our exact curriculum learning algorithm.
    \item Appendix \ref{app:hyperparams} contains the hyperparameters that were used for each network.
    \item Appendix \ref{app:compute} contains more details on the exact compute hardware used for training.
    \item Appendix \ref{app:other_results} contains additional experimental result tables and figures, including optimality gap scores, scores for all training conditions, performance breakdown by task, and learning curves.
    \item Appendix \ref{app:scoring} contains more details on how scores were calculated.
    \item Appendix \ref{app:performance} contains more details about the exact simulator performance under different conditions.
\end{itemize}

Our code is is publicly available, and links to the code and data can be found at at \url{https://generallyintelligent.com/avalon}.

\section{Game details}\label{app:biomes_objects_etc}
\subsection{Terrain, biomes, scenery, objects}

The base terrain is randomly generated via an iterative process of subdivision, bicubic interpolation, and adding various types of random noise at progressively smaller scales (see the \verb|build_outdoor_world_map| function in the code for an exact specification). This heuristic approach is designed to mimic the rugged surfaces of natural landscapes and mountains without using a significant amount of compute. A plane of water intersects the island at a specified height to form the surrounding body of water as well as pools within.

The terrain slope, elevation, and distance from water map every point to a set of 7 visually distinct biomes: \textit{tropical rain forest, tropical seasonal forest, temperate rain forest, temperate deciduous forest, grassland, temperate desert,} and \textit{subtropical desert}. Additional logic shapes regions around water into more specialized biomes such as beaches and swamps (\textit{water, fresh water, coastal, swamp, dark shore}) and adds a few bigger mountains and hills (whose surfaces are either \textit{bare}, in which case they can be climbed, or \textit{unclimbable}, in which case they cannot) to the landscape. We have tuned the resulting distributions to get worlds that feel both natural and high-variance. The island is then populated with foliage in a biome-dependent manner, with different levels of noise on the borders and density variations that create clearings for the player or agent to walk through.

Scenery elements include four types of trees, a bush, a flower, and a mushroom. Their arrangement and density are biome-specific --- for example, tropical biomes have mostly palm trees. Each piece of scenery populates specific biomes at specific spatial densities, with additional logic at biome boundaries and near coasts. All scenery objects have noise applied to their colors, scales, and orientations. Trees are climbable so that players can use them to escape predators or gain a better vantage point, while other smaller scenery items are non-colliding (agents can simply walk through them). In addition to scenery objects, a series of interactable items are scattered about in different locations and densities depending on the task. These include logs, sticks, large heavy boulders, medium-sized stackable stones, and fist-sized rocks. Many tasks require that these items be either maneuvered into new positions or used as tools or weapons. 

\begin{figure}[ht]
  \centering
    \includegraphics[width=1.0\linewidth]{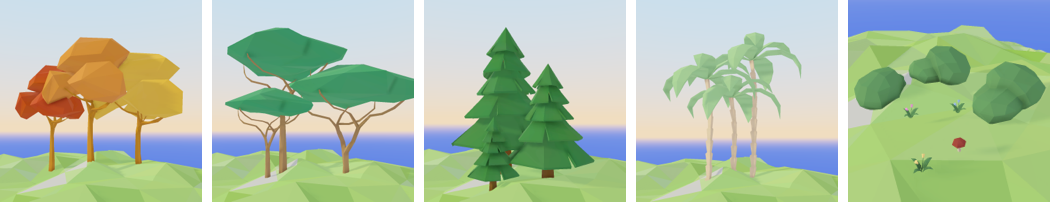}
    \caption{\textbf{
    Scenery objects.}
    Left to right: maple trees, acacia trees, fir trees, palm trees, and a composite shot that includes bushes, flowers, and a mushroom.
    }
    \label{fig:scenery}
\end{figure}

\begin{figure}[ht]
  \centering
    \includegraphics[width=1.0\linewidth]{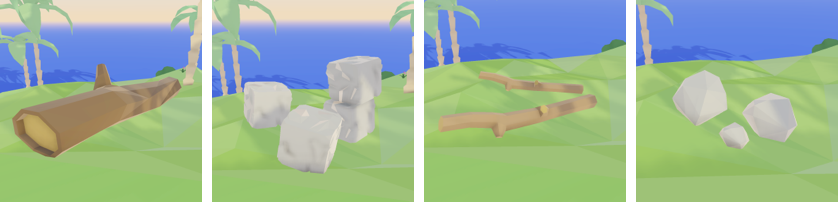}
    \caption{\textbf{
    Interactable items.}
    Left to right: log, boulders, sticks, and rocks.
    }
    \label{fig:items}
\end{figure}

\subsection{Obstacles} 

To procedurally enforce obstacles between the player and food, the terrain is shaped into irregular rings encircling either the player spawn point or the food location. These rings can consist of cliffs, chasms, ridges, or sudden drops that make it impossible for the player to get to the food without accomplishing the relevant task. Compositional tasks use concentric rings to impose a series of obstacles between the player and food. Noise is added to these rings to make them resemble more natural formations.

\subsection{Fruit trees and fruits}

\begin{figure}[h!]
  \centering
    \includegraphics[width=0.6\linewidth]{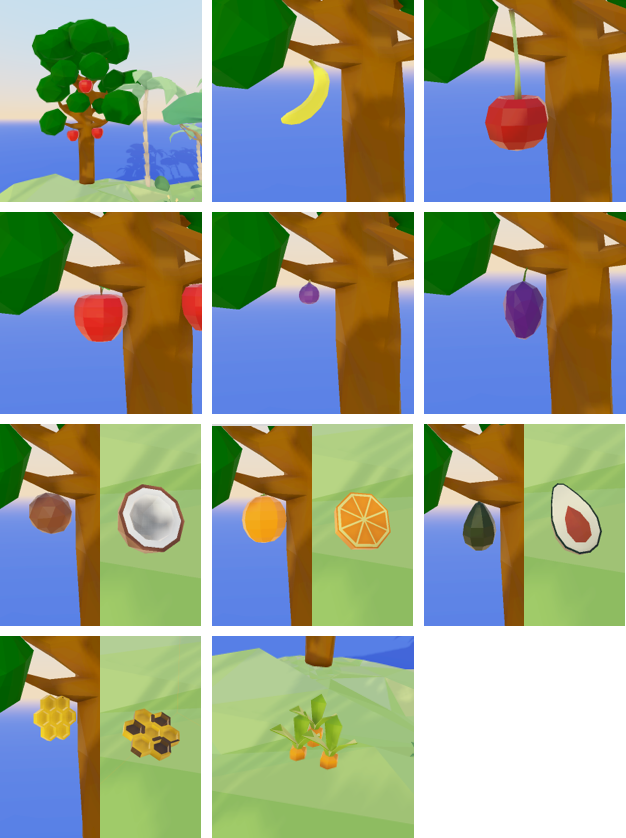}
    \caption{\textbf{Fruit tree and fruits}. Top left to bottom right: fruit tree, banana, cherry, apple, mulberry, fig, coconut (left: on tree, right: opened), orange (left: on tree, right: opened), avocado (left: on tree, right: opened), honeycomb (left: on tree, right: dirty), carrots. 
    }
    \label{fig:fruits}
\end{figure}

For most individual tasks, the available food item is a fruit that can be found either on or under a large fruit tree. Fruit trees are taller and more visually striking than other foliage (see Figure~\ref{fig:fruits}) so that identifying distant food is not akin to spotting a needle in a haystack.

The canonical fruit is an apple, but there are nine others that differ from it in some key aspect. These were selected not only for visual variety but to challenge the agent in different ways; for example, some require special preparation to become edible, while others are delicate and can become inedible with the wrong treatment. The full list of fruits and their properties is provided in Table~\ref{tab:fruits} (the word ``fruit'' is used loosely here, as it also includes honeycombs and avocadso). There are also carrots, which are the only food that can grow anywhere i.e. not restricted to be found under a tree or in a building.

The only tasks that feature a fruit other than the canonical apple are \texttt{eat}, which has a single fruit that can be any of the 10, and the compositional task \texttt{survive}, which has multiple different types of fruits scattered around the island.

\begin{table}[h!]
  \caption{Fruits and their properties.}
  \label{tab:fruits}
  \centering
  \begin{tabular}{ll}
    \toprule
    Fruit     & Description     \\
    \midrule
    Apple       & Canonical fruit.  \\
    Banana      & Provides less energy.  \\
    Fig         & Inedible (splattered) if it experiences too much force.  \\
    Orange      & Must be opened by making it experience a force.  \\
    Avocado     & Must be opened by hitting with a rock.  \\
    Coconut     & Must be opened by dropping from a height.  \\
    Honeycomb   & Inedible (dirty) if it contacts the ground.  \\
    Cherry      & Dark apple with long stem, always multiple.  \\
    Mulberry    & Smaller than other fruit.  \\
    Carrot      & Must be pulled from the ground by its leaves. \\
    \bottomrule
  \end{tabular}
\end{table}

\subsection{Buildings and doors} 

In addition to the natural features of the island, Avalon also includes procedurally-generated buildings which can be entered and explored. Buildings are used as an alternative site for placing fruits and can serve as a site for some tasks such as \texttt{open}, \texttt{push}, \texttt{navigate} and \texttt{stack}. Buildings may have one or more stories and parts of the buildings may be climbable. They are also guaranteed to contain at least one piece of food.  The \texttt{open} task takes place exclusively in buildings, with a range of door types and locks corresponding to different difficulties of the task. Doors can be unlocked or locked, and rotating (so the agent can push through it), or sliding (so the agent must drag it to the side). Doors can have between zero and three locks. There are three types of locks: a deadbolt that must be slid to the side, a rotating bolt that must be rotated to the side, and a timed button switch that toggles the ability to move the locks or open the door.

\subsection{Predators and prey}

\begin{figure}[h!]
  \centering
    \includegraphics[width=0.6\linewidth]{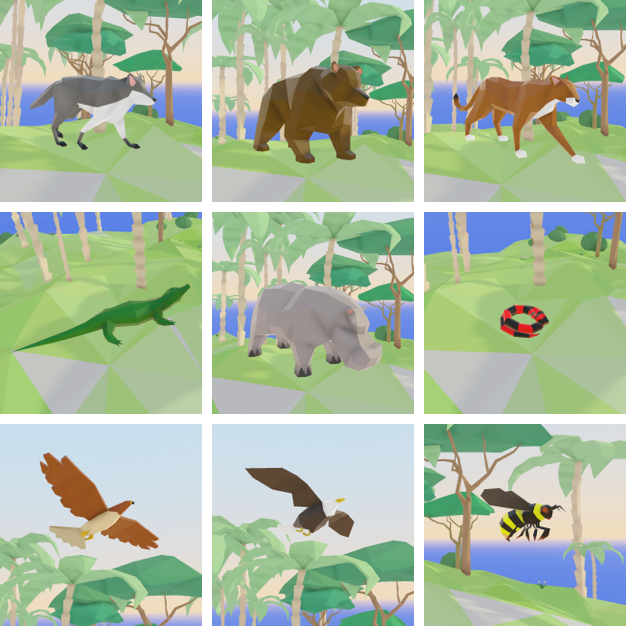}
    \caption{\textbf{
    Predators.}
    Top left to bottom right: wolf, bear, jaguar, crocodile, hippo, snake, hawk, eagle, bee.
    }
    \label{fig:predators}
\end{figure}

\begin{figure}[h!]
  \centering
    \includegraphics[width=0.8\linewidth]{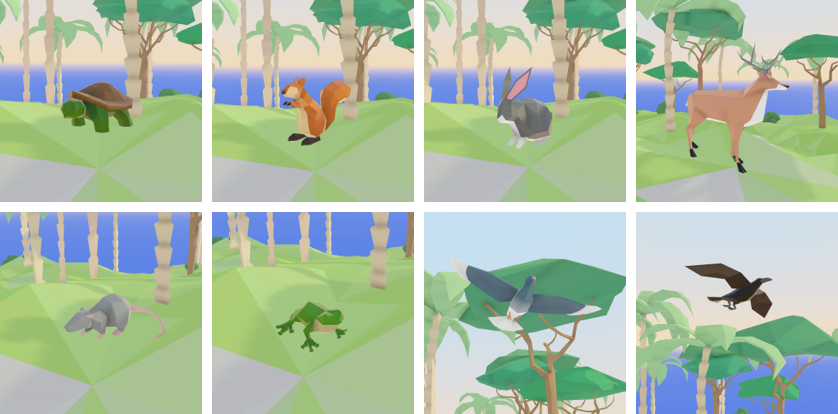}
    \caption{\textbf{
    Prey.}
    Top left to bottom right: turtle, squirrel, rabbit, deer, mouse, frog, pigeon, crow.
    }
    \label{fig:prey}
\end{figure}

There are 17 animals (nine predators, eight prey) that appear in certain tasks. Animals were selected to span a wide range of behaviors and game mechanics, described in detail below. Depending on the task, animals are either populated randomly or clustered near the food. The full list of animals and their properties is provided in Tables~\ref{tab:predators} and~\ref{tab:prey} and screenshots are provided in Figures~\ref{fig:predators} and~\ref{fig:prey}. In the remainder of this section, we briefly describe the axes of variation.

\begin{table}[h!]\tiny
  \caption{Predators (ordered from easiest to hardest) and their properties.}
  \label{tab:predators}
  {\centering
  \begin{tabular}{cccccccc}
    \toprule
    \textbf{Animal} & \textbf{Domain} & \textbf{Activation} & \textbf{Deactivation} & \textbf{Idle Behavior} & \textbf{Active Speed} & \textbf{Attack} & \textbf{Defense}    \\
    \toprule
    \dw{Bee} & \dw{Air} & \dw{Within Radius (Small)} & \dw{Outside Radius, KO} & Territory Bound, & \dw{Fast} & $-0.25$ & \dw{$+1$} \\
    & & & & Fast Random & & Perm. Respite & \\
    \midrule
    Snake & Ground & Within Radius (Small) & KO & Static Avoidant & Fast & $-5.0$ & $+1$ \\
    \midrule
    \dw{Hawk} & \dw{Air} & Within Radius (Wide) & Outside Radius & Territory Bound, & \dw{Fast} & $-0.25$ & \dw{$+1$} \\
    & & Within Territory & Outside Territory, KO & Periodic Fixed & & Temp. Respite & \\
    \midrule 
    \dw{Hippo} & \dw{Ground, No Inside} & \dw{Within Radius (Wide)} & Outside Radius, & \dw{Static Avoidant} & \dw{Fast} & $-0.5$ & \dw{$+\infty$}\\
    & & & Outside Domain & & & Persistent & \\
    \midrule 
    \dw{Alligator} & \dw{Ground} & \dw{Within Radius (Wide)} & Outside Radius, & \dw{Slow Random} & \dw{Slow} & $-0.5$ & \dw{$+1$} \\
     & & & Outside Domain, KO & & & Persistent & \\
    \midrule 
    \dw{Eagle} & \dw{Air} & Within Radius (Wide), & \dw{You See It, KO} & \dw{Fast Random} & \dw{Fast} & $-0.5$ & \dw{$+1$} \\
    & & You Don't See It & & & & Persistent & \\
    \midrule 
    \dw{Wolf} & \dw{Ground} & Within Radius (Wide), & Outside Radius, & Territory Bound & \dw{Fast} & $-0.5$ & \dw{$+2$} \\
    & & Within Territory & Outside Domain, KO & Slow Random & & Persistent &  \\
    \midrule 
    \dw{Jaguar} & \dw{Ground, Climb} & Within Radius (Wide), & Outside Radius, & \dw{Periodic Fixed} & \dw{Fast} & $-0.5$ & \dw{$+2$}\\
    & & Must See You & You Stop Moving, KO & & & Persistent & \\
    \midrule 
    \dw{Bear} & Ground, Climb, & \dw{Within Radius (Wide)} & Outside Radius, & \dw{Slow Random} & \dw{Fast} & $-0.5$ & \dw{$+\infty$} \\
    & No Inside & &  Outside Domain & & & Persistent & \\
    \bottomrule
  \end{tabular}}
\end{table}

\begin{table}[h!]\scriptsize
  \caption{Prey (ordered from easiest to hardest) and their properties.}
  \label{tab:prey}
  {\centering
  \begin{tabular}{ccccccc}
    \toprule
    \textbf{Animal} & \textbf{Domain} & \textbf{Activation} & \textbf{Deactivation} & \textbf{Idle Behavior} & \textbf{Active Speed} & \textbf{Defense}    \\
    \toprule
    Frog & Ground & n/a & KO & Slow Random & n/a & $+1$ \\ 
    \midrule
    Turtle & Ground & Within Radius (Small) & Outside Radius, KO & Periodic Fixed & Slow & $+1$ \\
    \midrule
    \dw{Mouse} & \dw{Ground, Climb} & Within Radius (Small), & Outside Radius, & \dw{Fast Random} & \dw{Fast} & \dw{$+1$} \\
    & & Must See You & Can't See You, KO & & & \\
    \midrule
    Rabbit & Ground & Within Radius (Wide) & Outside Radius, KO & Fast Random & Fast & $+1$ \\
    \midrule
    \dw{Pigeon} & \dw{Air} & \dw{Within Radius (Small)} & \dw{Outside Radius, KO} & Territory Bound, & \dw{Fast} & \dw{$+1$} \\
    & & & & Fast Random & & \\
    \midrule
    Squirrel & Ground, Climb & Within Radius (Small) & Outside Radius, KO & Fast Random & Fast & $+1$ \\
    \midrule
    Crow & Air & Within Radius (Wide) & Outside Radius, KO & Slow Random & Fast & $+1$ \\
    \midrule
    Deer & Ground & Within Radius (Wide) & Outside Radius, KO & Static Avoidant & Fast & $+2$ \\
    \bottomrule
  \end{tabular}}
\end{table}

\textbf{Domain.} Animals are either confined to the ground or exclusively fly. Some ground animals can climb. All animals can enter buildings except bears and hippos, which are too big to fit through doors. 

\textbf{Activation and deactivation.} At any given moment, an animal is either \textit{active} or \textit{inactive}, with inactivity the default. Different predators and prey have different activation (deactivation) conditions for when they start (stop) chasing or fleeing the player, respectively. For example, some only activate when the player is nearly on top of them while others activate from a wider radius. Activation conditions are often mirrored in deactivation conditions, though this is not always the case; for example, hawks and wolves both activate when the player enters their territory, but wolves continue the pursuit even after the player exits their territory. Most predators and all prey can be deactivated by killing, but two predators are indestructible (hippo, bear). Another way to deactivate some predators is by triggering the Outside Domain condition. For example, predators that can neither fly nor climb can be deactivated by climbing a tree. Hippos and bears are too big to fit through doors so going inside a building triggers this condition for them. Some special activation conditions: "Must See You" means the player must be within the predator's field of view and "You Don't See It" means it sneakily attacks when the player is not looking. For deactivation conditions, "You See It" is the mirror of "You Don't See It" and "You Stop Moving" is akin to the freeze defense strategy for not getting eaten.

\textbf{Idle behavior.} While inactive, animals display different default idle behaviors. Some move randomly (fast or slow) while others move along fixed periodic trajectories. Some only prowl within their territory. Static Avoidant animals are static until the player approaches, then slowly move away to increase their distance from the player, then finally activate if the player get within their activation threshold.

\textbf{Active speed, attack and defense stats.} When active, some animals are slower than the player, while others are faster. Some predators attack repeatedly (Persistent) while others attack once and then leave for a bit (Temporary Respite). Bees attack once and then die (Permanent Respite). Attacks can be high damage or low damage. A single snake attack is fatal. Some animals can be disabled with one hit, others need multiple hits, and some are indestructible.

\section{Reward function}\label{app:reward}
In all tasks, the agent's reward is given by the same reward function of its energy level over time.
Agents begin each episode with a certain initial energy and must inevitably expend energy in moving to complete the task.
The only way to gain energy is by eating.
In addition to energy expenditure from motion, energy can be lost by taking damage from enemies or falls. Episodes terminate for one of three reasons:
\begin{enumerate}[i)]
    \item The agent energy goes to zero.
    \item The agent eats all the food in the episode (the episode ends 10 frames later).
    \item The episode times out.
\end{enumerate}
The determination of the initial energy, definition of energy expenditure, and implementation of termination conditions vary slightly between training and evaluation settings, described in Section~\ref{sec:rewards:train_eval}. The formulas for each energy term and values of associated constants are provided in Section~\ref{sec:rewards:energy}.

\subsection{Training and evaluation}\label{sec:rewards:train_eval}

During training, the agent receives dense reward equal to its framewise change in energy level, together with framewise penalties. This reward $R$ is the sum of any energy gained from eating $E_{\rm food}$ minus any energy lost from damage due to falling $E_{\rm fall}$ and predators $E_{\rm predators}$, minus the movement penalty $P_{\rm move}$ and frame penalty $P_{\rm frame}$,

\begin{equation}\label{eqn:reward}
    R=E_{\rm food} - E_{\rm fall} - E_{\rm predators}  - P_{\rm move}- P_{\rm frame}
\end{equation}

Each energy term is discussed in detail in Section~\ref{sec:rewards:energy}. During training and evaluation, the initial energy level $E_0=1$. The time limit during training depends on the difficulty level and task, tuned such that 98\% of humans trials lie within the time limit. These time limits improve compute efficiency during training, terminating episodes in which the agent cannot perform the task.

\subsection{Energy terms}\label{sec:rewards:energy}

Each of the energy terms in Equation~\ref{eqn:reward} are described below, with all constants listed in Table~\ref{tab:reward_constants}.
 
\textbf{Food}. Food is eaten once it is held and within 0.5m of the head. All food (fruits and prey) provides $E_{\rm food} = 1$ upon being eaten, except for bananas which provide $E_{\rm food} = 0.5$ (see Table~\ref{tab:fruits}).

\textbf{Falling}. Fall damage occurs if the vertical speed of the agent $v_{\rm vertical}$ is above a certain threshold $v_{\rm threshold}$ when colliding with the ground, with the energy penalty from a fall given by

\begin{equation}
    E_{\rm fall} =  C_{\rm fall}m_{\rm body}\max\left(0,  v_{\rm vertical}^2 - v_{{\rm threshold}}^2\right)
\end{equation}

The energy coefficient $C_{\rm fall}$ and threshold $v_{\rm threshold}$ are set so the agent of body mass $m_{\rm body}$ would not take damage from any fall smaller than 12 meters and take lethal damage when falling from anything greater than 28 meters.  See Table~\ref{tab:reward_constants} for details.

\textbf{Predators}. The energy cost of predator attacks depends on the predator as specified in Table~\ref{tab:predators}.

\subsection{Penalty terms}\label{sec:rewards:movement}

The penalty terms differ from the energy terms in that they do not modify the total agent energy: they do not modify the ability of the agent to survive or act in its environment. Rather, these terms are designed to assist the agent in choosing more appropriate actions which could improve learning and generalization. As such, they contain a number of parameters that may be tuned to optimize agent performance.

\textbf{Frame}. Our scoring criteria (Section~\ref{app:scoring}) adds a bonus if all the food on a level is consumed before the end of the episode. This bonus is proportional to the number of frames remaining before the timer would terminate the episode. We use the same proportionality constant used for scoring (1e-4) as our frame penalty $P_{\rm frame}$.

\textbf{Movement}. As stated in Section~\ref{sec:interface}, the agent controls its head, body, and two hands. The head and body are generally controlled simultaneously. Translating the head in any direction except vertically translates the entire agent. Similarly, rotating the head about the vertical axis (i.e. looking from side to side) reorients the entire agent. However, when it comes to the vertical axis, the agent's head and body are treated as distinct. When translating the head vertically, only the head moves; this is to enable crouching, during which the agent is still resting on the ground but its vantage point is lower. To vertically translate the agent's body as well, such that its distance from the ground changes, the agent must jump, climb, or fall. Similarly, rotating the head about axes other than the vertical axis only reorients the head, not the body. This is so that the agent can look up and down or tilt its head to look at things without doing somersaults. 

With this formalism in place, the energy cost of movement can be stated as the sum of kinetic $\Delta K$ and gravitational potential $\Delta P$ energy changes for each of the agent's head, body, and two hands,

\begin{equation}
    P_{\rm move} = C_{\rm move}\left( \sum_{\substack{i \in [\rm{body}, \\ {\rm hands}]}} (\Delta K_{i} + \Delta P_{i}) + \Delta P_{\rm head} \right),
\end{equation}

where the energy costs of the body capture all global translations and rotations of the full agent and the energy costs of the head and hands are calculated relative to the body. For the head, there is only a potential energy cost to penalize keeping the head tilted; there is no kinetic energy cost for crouching. The overall movement coefficient $C_{\rm move}$ and the kinetic and potential energy terms are determined by the physics-based formulas,

\begin{align}
        \Delta K_i &= C_{\mathrm{K}i}m_i \max\left(0,\,\bar{\mathbf{v}} \cdot \Delta \mathbf{v}\right), \\
        \Delta P_i &=C_{\mathrm{P}i} m_i g \max\left(0,\,\Delta h\right),
\end{align}

where $C_{\mathrm{K}i}$ and $C_{\mathrm{P}i}$ are coefficients that can be tuned, $\bar{\mathbf{v}}$ is the average velocity over the last step, $\Delta \mathbf{v}$ is the change in velocity, $m_i$ is the mass of the part $i$, $g$ is the gravitational acceleration, and $\Delta h$ is the change in height over the last step. The constants used to calculate these energy terms are listed in Table~\ref{tab:reward_constants}. During training, $C_{\rm move}$ was tuned with other hyperparameters. The body part masses were set to be close to the average human equivalents.

\begin{table}[h!]
  \caption{Parameters used to calculate fall and movement energy values. $C_{\rm move}$ is a hyperparameter that can be adjusted for training. During evaluation, $C_{\rm move}=0$}.
  \label{tab:reward_constants}
  \centering
  \begin{tabular}{ll}
    \toprule
       Parameter   &   Value     \\
    \midrule
       $C_{\rm fall}$      & 0.000026 \\
       $v_{\rm threshold}$ (m/s) & 10 \\
       $C_{\rm move}$         & 1e-8 \\
       $P_{\rm frame}$         & 1e-4 \\
        $m_{\rm body}$ (kg)     &  60 \\
        $m_{\rm head}$ (kg)     &  4.8 \\
        $m_{\rm hand}$ (kg)     &  3.0 \\
        $g$ (${\rm m}/{\rm s}^2$)     &  10 \\
       $C_{\rm K}$         & 0 \\
       $C_{\rm P}$         & 1 \\
    \bottomrule
  \end{tabular}
\end{table}





\section{Observation space}\label{app:observation}
The agent's observation space is made up of $96 \times 96 \times 4$ egocentric RGB + depth images and the following 23 proprioceptive inputs:

\begin{itemize}
    \item $4 \times 2$ variables (3D vectors) corresponding to the framewise change in position and rotation of the agent's body, head, and two hands.
    \item $3 \times 2$ variables (3D vectors) corresponding to the position and  rotation of the agent's head and two hands relative to its body.
    \item $2 \times 1$ variables (booleans) corresponding to whether each hand is colliding with anything.
    \item $2 \times 1$ variables (booleans) corresponding to whether each hand is grasping something.
    \item 4 variables (floats) corresponding to the agent's energy expenditure from movement, energy lost from enemy attacks, energy gained from eating, and current energy level.
    \item  1 variable (float) corresponding to the frames remaining before the episode is terminated by the timer.
\end{itemize}

\section{Action space}\label{app:action}
The full 21-dimensional action space includes $3 \times (3 + 3)$ continuous degrees of freedom for the 3D translation and rotation of the agent's three components (head and two hands), as well as $2 \times 1$ binary grasp actions for the hands and $1$ binary jump action. This action space roughly corresponds to the controls of a VR headset and was used for all experiments in the paper.  Human players were also allowed to use the controller joystick for movement and rotation, but these inputs were carefully mixed into the simpler action space in which the agent acts so that the two could be exactly equivalent.

We also provide a reduced 9-dimensional action space which maps to mouse and keyboard controls. In this reduced action space, there is a single translational degree of freedom for forward/backward motion, with the agent always moving in the direction it is facing, which is controlled by two angular degrees of freedom (pitch and yaw). In addition to the $2 \times 1$ binary grasp actions and 1 continuous jump action, this setting also has 1 discrete action for eat and $2 \times 1$ discrete actions for throw.

\section{Factors of variation}\label{app:FOV}
The procedure that generates Avalon's levels is parameterized by many \textit{factors of variation} which determine the structure and appearance of the generated world.
These factors of variation can be broken down into settings that affect terrain (shape and color), scenery (trees, bushes, flowers, etc), environment (sky, lighting and graphics options), buildings, items (animals, tools, and food), and tasks (distributions over factors like how far to jump, how many enemies to include, etc).
Two levels generated from the same factors of variation will share these high-level features but will differ in various low-level details like object positions and the exact topography of the island.

Due to the focus on variety and diversity in Avalon, the factors of variation are mostly defined in code rather than in external configuration files. This also allows for better specification of allowed values (via type signatures).

\subsection{Terrain}

The terrain (base world geometry) is generated via repeated subdivision and addition of various types of noise. The \verb|WorldConfig| object defines 34 properties that control this procedure, including 
\verb|fractal_iteration_count|, 
\verb|noise_scale_decay|, and 
\verb|size_in_meters|.
See the code for a complete list and documentation for each value.

The \verb|generate_world_config|
function gives an example of how to dynamically generate interesting, varied \verb|WorldConfig|s. The \verb|build_outdoor_world_map| converts a \verb|WorldConfig| into a \verb|HeightMap|. It can easily be swapped out for any other approach to generating a \verb|HeightMap| (grid of heights).

After generating the base terrain, the world is broken into different ``biomes" that control which scenery objects will be placed in that region, as well as the colors used for the terrain in that region. Some of these biomes affect the height of the world (ex: the beach biome controls the erosion of shores near the ocean). This entire process of biome assignment and calculation is controlled by the \verb|generate_biome_config| function, which generates a \verb|BiomeConfig|. This object has 43 properties such as \verb|beach_slope_cutoff|, \verb|swamp_elevation_max|, and \verb|rock_color_noise|. See the code for a fully documented list.

\subsection{Scenery}

In order to strike a good balance between visual complexity and performance, biomes in Avalon are populated with instanced scenery models. These models are given per-instance and per-vertex color noise, as well as per instance scaling and rotation variation without really incurring any significant performance overhead. These attributes are controlled by the \verb|FloraConfig| and \verb|SceneryConfig| objects, which together define 10 attributes like \verb|density|, \verb|border_mode|, and \verb|correlated_scale_range|.

\subsection{Environment}

All aspects of the Godot \verb|WorldEnvironment| and \verb|Sky| objects are procedurally generated, as well as all aspects of the \verb|Sun| light that is created in each scene. This enables varying factors such as \verb|sun_latitude|, \verb|fog_color| and \verb|tonemap_mode|. We give examples of setting 46 of these, though all supported properties of these objects in Godot can be set directly. See the Godot documentation for more details on each setting.

\subsection{Buildings}

Buildings are used both as components of compositional worlds and as variants of the basic tasks (ex: \texttt{explore} tasks are set either inside of a building or in an outdoor, natural world).  The appearance of buildings can be changed via the \verb|BuildingAestheticsConfig|, which contains 20 attributes like \verb|desired_story_count|, \verb|window_width| and \verb|trim_color|

\subsection{Items}

All items, including animals, have a variety of attributes which can be set directly and differ per-item. See \verb|items.py| for a complete definition of all attributes. Each item also has \verb|safe_scale| and \verb|base_color| attributes, which can be used to explicitly set the size and color of each object. Animals each have a variety of behavior-dependent attributes like speed, activation radius, etc that can be altered for each instance.

\subsection{Tasks}

Tasks are each generated by a single function, parameterized only by difficulty. This function converts that difficulty value into all other lower-level factors of variation (e.g. jump distance for \texttt{jump}, path width for \texttt{move}, etc). See the corresponding \verb|.py| files in the \verb|tasks| folder for a complete specification of each task's parameters. Tasks which can be composed also contain a single function to create the given type of obstacle (ex: a gap for jumping over, a path for walking along, etc), and these functions are used by \verb|compositional.py| to implement the compositional tasks.

\section{Task definitions}\label{app:tasks}
\begin{figure}[h!]
  \centering
    \includegraphics[width=0.9\linewidth]{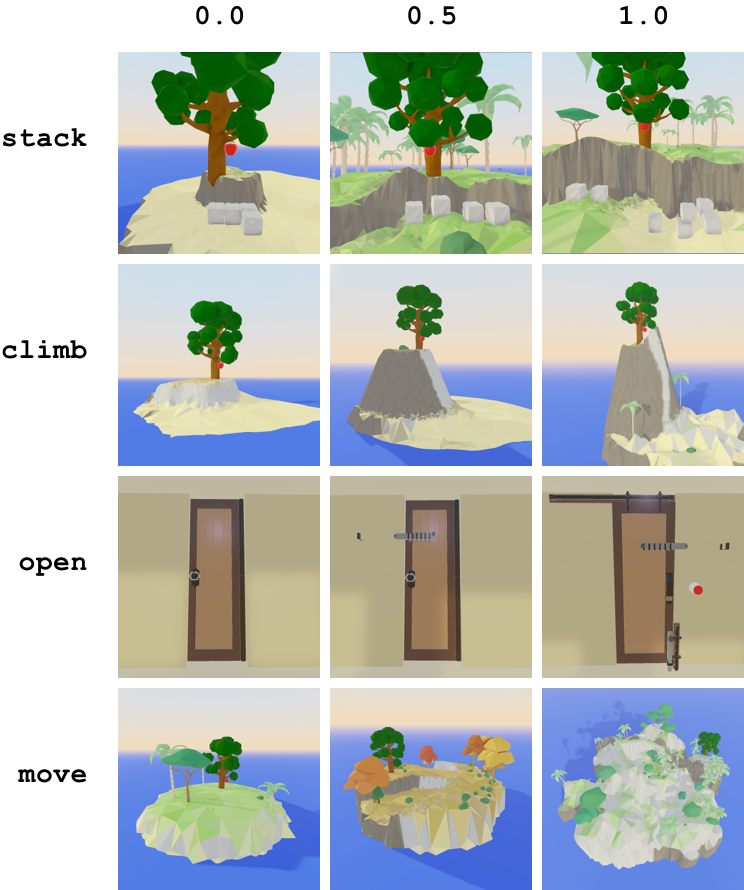}
    \caption{\textbf{
    Task variation with difficulty.}
    Examples of how tasks vary as difficulty increases. \texttt{stack} $0.0$ is already solved and the agent just needs to jump on the blocks to reach the landing; \texttt{stack} $1.0$ requires the agent to create a pyramid of blocks that is three blocks high in order to reach the landing. \texttt{climb} $1.0$ requires climbing a higher cliff, on a narrower path than \texttt{climb} $0.0$. In \texttt{open} $0.0$ the agent can directly walk through the door; \texttt{open} $1.0$ requires the agent to unlock three locks, including a timed switch, to open the door. \texttt{move} $0.0$ requires moving a short distance on flat land to get food, whereas \texttt{move} $1.0$ requires moving a longer distance on a narrow path surrounded by cliffs.
    }
    \label{fig:difficulty}
\end{figure}

Each task in Avalon maps to a set of generated worlds. Each world is set up such that the agent must usually complete the task in order to successfully reach the food. Each task has a variety of individual parameters, controlled in aggregate by the difficulty parameter, which ranges from $0$ to $1$. See Figure~\ref{fig:difficulty} for examples of how tasks vary by difficulty.

Almost every task world has only one fruit or prey; after the agent eats the fruit or prey, the episode ends. The two exceptions are \texttt{gather}, which has multiple fruits, and \texttt{survive}, which has multiple prey and fruits. To enable visibility for the agent in rugged terrain, the fruit in each world is always on a fruit tree or inside a building.

Each tasks is set up with the goal of isolating the skill being learned. Thus, all tasks only have apples as canonical fruit at all difficulty levels, with the exception of more difficult \texttt{eat} worlds, which contain harder-to-eat fruits; \texttt{hunt} and \texttt{throw}, which contain only prey and no fruit; and \texttt{survive}, which can contain all fruit and prey types. This is intended to isolate the skills needed for each task from the skills needed to eat more complex fruits. Additionally, in most tasks, with the exception of \texttt{explore}, \texttt{find}, \texttt{gather}, and \texttt{survive}, the agent and the fruit spawn in locations such that the agent can see where the fruit is at spawn. In this case, the agent may not be facing the fruit at spawn, but upon turning in place will be able to see the fruit. This is intended to isolate the skills required for each task from the skills required for exploring the world, which is tested in isolation in \texttt{explore}, as well as in the compositional tasks.

\subsection{Basic tasks}

There are 16 "basic" tasks. Each basic task generates a world in which the agent usually must complete that task in order to reach and eat the food.

\begin{figure}[h!]
  \centering
    \includegraphics[width=1.0\linewidth]{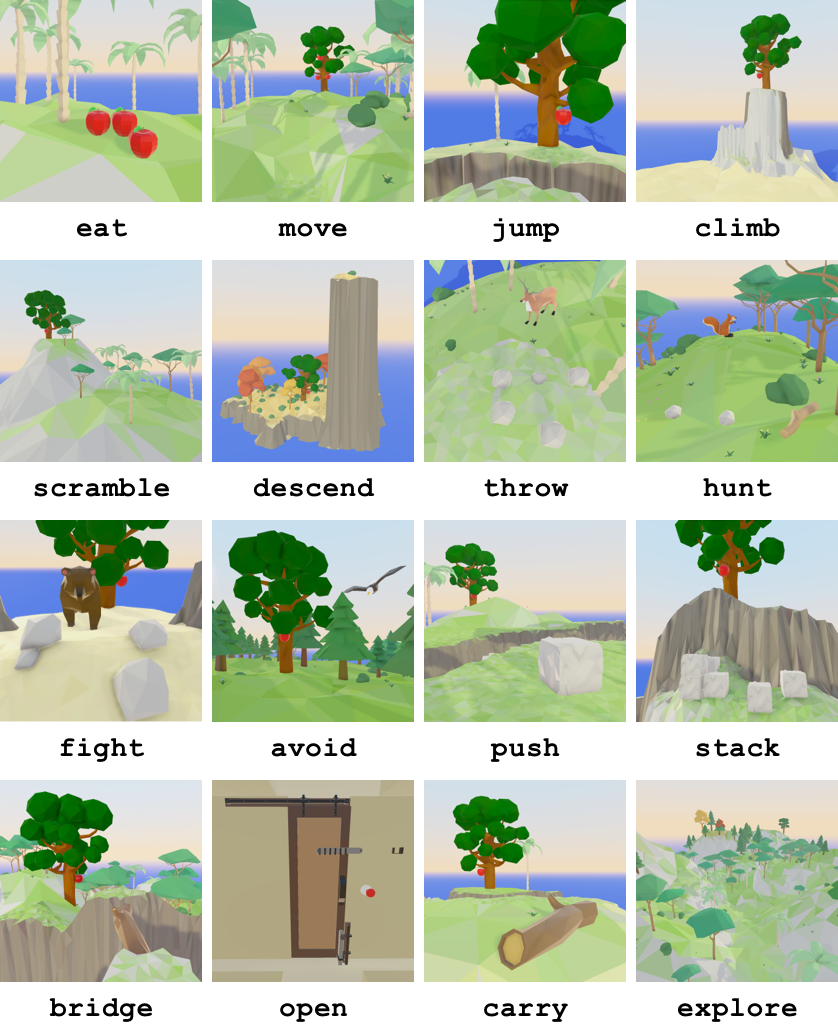}
    \caption{\textbf{
    Basic tasks.}
    All 16 basic tasks. For each task, the world is generated such that the agent must complete the task in order reach and eat the food.
    }
    \label{fig:basic}
\end{figure}

\textbf{Eat}.
In \texttt{eat}, the agent starts in front of the fruit, and must grab the fruit and bring the fruit to within a radius of its head in order to succeed.
As difficulty increases, worlds contain different types of fruit that must be opened in more complex ways (see Appendix~\ref{app:biomes_objects_etc} for details on fruit types). 

\textbf{Move}.
In \texttt{move}, the agent starts a distance away from the fruit, and must move its body to reach the fruit. 
As difficulty increases, the agent spawns farther away from the fruit, and the paths leading to the fruit become narrower. At the highest difficulties, the agent must traverse thin, multi-segment paths across chasms in order to reach the fruit.

\textbf{Jump}.
In \texttt{jump}, the agent must jump over a chasm to reach the fruit. 
As difficulty increases, the chasm gets wider, and the area that can be jumped across gets narrower, such that the agent must figure out where it can successfully land a jump. If agents fall into the chasm while jumping, they can climb back out to the original side and retry the jump.

\textbf{Climb}.
In \texttt{climb}, the agent must climb up a cliff wall in order to reach the fruit. The climbing motion requires repeatedly bringing one hand up, grabbing the cliff, pulling the hand downward, and then doing the same with the other hand, without letting go of the cliff. 
As difficulty increases, the cliff gets taller so the agent must climb further, and the climbable path gets narrower so the agent has less area to grab onto. The climbable path also contains multiple, angled segments at higher difficulties, requiring more than just climbing in a single straight line.

\textbf{Scramble}.
In \texttt{scramble}, the agent must combine walking, jumping and climbing to move over terrain to reach the fruit. 
As difficulty increases, the terrain becomes more mountainous, requiring more climbing and less jumping.

\textbf{Descend}.
In \texttt{descend}, the agent must descend down a cliff. This can be accomplished by climbing down, or falling while grabbing the cliff wall periodically. On easier difficulties, there is also a platform partway down that can be used to safely descend by first falling on the platform, then falling the rest of the way. If the agent just jumps off the cliff randomly (without falling on this platform), it will take damage.
As difficulty increases, the cliff gets taller so the agent will take more damage (and, at the highest difficulties, die). The path that can be grabbed also gets narrower, so the agent must figure out where to descend from.

\textbf{Throw}.
In \texttt{throw}, the agent must throw a rock at prey in order to kill it, and then eat it in the same way fruit is eaten. \texttt{throw} worlds only have prey, and no fruit. 
As difficulty increases, the agent starts farther away from the prey, and the world gets larger, so the prey can run away. Additionally, the types of prey that spawn are harder to hit (e.g. pigeons that fly, instead of frogs that sit on the ground and hop slowly).

\textbf{Hunt}.
In \texttt{hunt}, the agent must find the prey and kill it either by throwing a rock, or hitting it with a stick. 
As difficulty increases, the world gets larger, the types of prey that spawn are harder to hit, and fewer tools spawn for hunting. The agent may only get one rock, or one stick.

\textbf{Fight}.
In \texttt{fight}, the agent must use sticks or stones to fight a predator that is guarding the food in order to reach and eat the food. While it is not strictly necessary to defeat the predators in order to reach the food in some levels, as difficulty increases, the predator types that spawn are more aggressive and more difficult to hit, and there are more of them, effectively forcing the agent to fight.

\textbf{Avoid}.
In \texttt{avoid}, the agent must avoid predators that are near the food in order to reach and eat it. Unlike in \texttt{fight}, the agent is not provided with sticks or stones for fighting the predator. 
As difficulty increases, the number of predators increase, and more aggressive types of predators become more likely to spawn.

\textbf{Push}.
In \texttt{push}, the agent must push a heavy boulder into a position where it can be used as a stepping stone for jumping onto the cliff ledge where the fruit is.
As difficulty increases, the boulder gets heavier and more difficult to control, and must be pushed farther. 
This task has both indoor and outdoor variants.

\textbf{Stack}.
In \texttt{stack}, the agent must stack stones and jump on top of them in order to reach the cliff ledge where the fruit is. 
At low difficulties, the agent only needs to use one stone to reach the top. As difficulty increases, the cliff ledge gets taller, so more layers of stones need to be stacked to reach the top. At the highest difficulty, the agent must stack a pyramid with four layers of stones in order to reach the top. This task has both indoor and outdoor variants.

\textbf{Bridge}.
In \texttt{bridge}, the agent must pick up a log and lay it across a chasm in order to create a bridge to walk over the chasm. 
At low difficulties, the log is sometimes already in the solved position. 
As difficulty increases, the log starts out farther away, and the width of the chasm section that is narrow enough to be bridged shrinks, so the agent must find the right place to bridge the chasm.

\textbf{Open}.
In \texttt{open}, the agent starts inside a building and must open a door to get into the room where the fruit is. See Appendix~\ref{app:biomes_objects_etc} for details on types of doors and locks. 
At the easiest levels, doors are rotating and can be walked through. As difficulty increases, more difficult variants of doors (such as sliding doors and doors that must be pulled rather than pushed) with more difficult locks (such as the timed switch) appear more often. At the highest difficulty, doors can all three locks.

\textbf{Carry}.
\texttt{carry} spawns a task world that requires objects from one of \texttt{throw}, \texttt{fight}, \texttt{stack}, or \texttt{bridge}, and then moves the objects a distance away from where they would normally spawn. Thus, the agent must carry the object to the location where it will be used.
As difficulty increases, objects spawn farther away and must be found and carried a longer distance.

\textbf{Explore}.
In all other basic tasks, fruit or prey is visible from where the agent spawns. In \texttt{explore}, fruit cannot be seen from where the agent spawns. Thus, the agent must explore the terrain in order to find the fruit.
This task has both indoor and outdoor variants.
In the outdoor variant, as difficulty increases, the world gets larger and the terrain rockier, making it more difficult to spot and reach the fruit.
In the indoor variant, the building to be explored becomes progressively larger with higher difficulties.

\subsection{Compositional tasks}

There are four "compositional" tasks. Compositional tasks generate worlds in which the agent must complete a sequence of multiple basic tasks in order to get food. To prevent the agent from bypassing any task, compositional tasks use concentric rings of terrain obstacles to impose a sequence of task obstacles, or spawn buildings that require one of the basic tasks.

\begin{figure}[h!]
  \centering
    \includegraphics[width=1.0\linewidth]{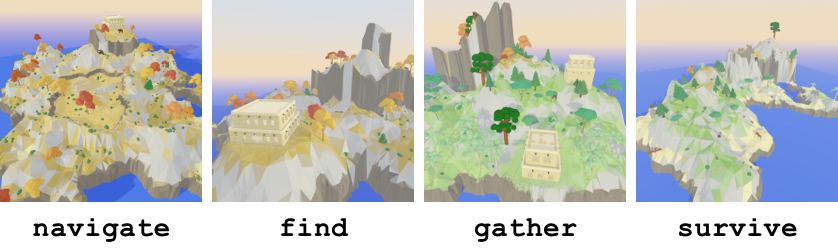}
    \caption{\textbf{
    Compositional tasks.}
    All four compositional tasks. Compositional tasks generate worlds in which the agent must complete a sequence of multiple basic tasks in order to get food.
    }
    \label{fig:compositional}
\end{figure}

\textbf{Navigate}.
In \texttt{navigate}, the fruit is visible from where the agent spawns, and the agent must navigate through a variety of basic tasks to reach the fruit. 
As difficulty increases, the world gets larger, with more difficult terrain, and the number and difficulty level of basic tasks that must be solved also increases (up to a maximum of four basic tasks).

\textbf{Find}.
\texttt{find} is like \texttt{navigate}, but the fruit is not visible from where the agent spawns. In order to achieve maximal score on the task, the agent must find the fruit and solve a variety of basic tasks to reach the fruit before time runs out.
As difficulty increases, the world gets larger and more mountainous, and the number and difficulty of basic tasks to be solved along the way also increases.

\textbf{Gather}.
\texttt{gather} is like \texttt{find}, but with multiple pieces of fruit in the world. None of the fruit is guaranteed to be visible from where the agent spawns. Thus, the agent must find and reach all pieces of fruit before time runs out, solving obstacles along the way, in order to achieve max score on the task.
As difficulty increases, the world gets larger and more mountainous, the distance between fruits increases, and the number and difficulty of basic tasks to be solved increases.

\textbf{Survive}.
\texttt{survive} is like \texttt{gather}, but with prey and predators scattered about the world, in addition to fruit and some basic task obstacles. Unlike in \texttt{navigate}, \texttt{find}, and \texttt{gather}, obstacles don't necessarily appear at all or need to be solved in sequence. Instead, the agent's goal is to survive as long as possible by eating prey or fruit, avoiding predators, and solving obstacles in order to find more food. 
As difficulty increases, the world gets larger and more mountainous, and there are more predators of higher difficulty and fewer prey and fruit. 
\texttt{survive} at the high difficulty levels is in some ways the pinnacle task: the agent usually needs to have learned a wide range of skills from all other tasks in order to achieve maximum scores on the most difficult survive levels.

\section{Evaluation worlds}\label{app:eval_worlds}
\subsection{World generation}

A fixed set of 50 worlds for each task was generated to evaluate human and agent performance. Each world was generated with a unique seed while difficulties ranged between 0.0 and 1.0 and the exact distribution of difficulties was task dependent. For each task type, 20\% of the worlds were set to have difficulty 1.0 and the remaining worlds had difficulties that were evenly spaced between 0.0 and 1.0. For \texttt{hunt}, \texttt{avoid} and \texttt{eat}, a world for each type of food, prey and predator were generated to guarantee that each type of entity was present in an evaluation world. For \texttt{avoid} and \texttt{hunt}, the difficulties of these forced worlds were set to 1.0, while for eat the difficulty was set to 0.5. The forced levels were created first and then the same procedure as the other tasks was used to generate the remaining worlds.

\subsection{Replaced worlds}

During data collection, participants reported worlds that were impossible or very difficult to complete. Of the original 1,000 levels generated (50 levels $\times$ 20 tasks), there were a total of 30 levels (3\%) that had any issues at all. Ultimately, 10 levels (1\%) were replaced.

Four levels were impacted by 2 bugs which have since been fixed. When these issues were encountered during data collection, we simply generated new worlds with an increased seed (and the same difficulty level) to replace them in the evaluation set.  Twenty-six levels did not have successful playthroughs in the initial round of data collection. These levels tended to be on the highest difficulty setting and most difficult tasks, and are listed in Table~\ref{tab:26_bad_worlds}. Of those 26 levels, we were able to eventually solve 20 by simply having more players try them. Of those 20 levels, 9 required multiple attempts from the same individual before they were ultimately solved; however, multiple attempts were needed only due to the difficulty of execution, not because they required advance knowledge of the level, and thus it seems likely to us that the level could have been solved on the first attempt given enough skilled players (but unfortunately we did not have enough study participants to verify). The remaining 6 did not get successful playthroughs even after additional attempts, and thus were replaced. However, only 1 of these 6 was truly impossible (the terrain is extremely jagged, leading to a situation where the fruit spawned in an area that is unreachable) whereas the others were merely very difficult.

In summary, 10 of the generated worlds (1\%) were replaced, but of those only 1 (0.1\%) was due to a level actually being impossible. We reiterate that the evaluation set does not contain any unsolvable worlds as they were all replaced, but we report these issues here for the sake of transparency.

\begin{table}[h!]
  \caption{List of 30 generated evaluation worlds for which there were initially no successful playthroughs. After inspection and additional playthroughs, 10 worlds were replaced in the evaluation set (see comments column), although only 1 was actually unsolvable after fixing bugs.}.
  \label{tab:26_bad_worlds}
  \centering
  \begin{tabular}{llll}
    \toprule
       Task & Seed & Difficulty & Comment     \\
    \midrule
    \texttt{eat} & 542140 & 0.5	  & Replaced (due to bug, now fixed).	 \\
    \texttt{eat} & 542145 & 0.5	  & Replaced (due to bug, now fixed).	 \\
    \texttt{avoid} & 542771 & 0.85  &     \\
    \texttt{avoid} & 542787 & 1	 &       \\
    \texttt{descend} & 542387 & 1 &  	 \\
    \texttt{fight} & 542731 & 1	  & Replaced (unlucky world, impossible). \\
    \texttt{fight} & 542694 & 1   &  	 \\
    \texttt{find} & 543024 & 0.92  &  	 \\
    \texttt{find} & 543031 & 1	  & Replaced (very difficult, no successful plays). 	 \\
    \texttt{find} & 543033 & 1	  & Replaced (very difficult, no successful plays).  	 \\
    \texttt{find} & 543034 & 1	  &  Replaced (very difficult, no successful plays). 	 \\
    \texttt{find} & 543017 & 0.74     &   \\
    \texttt{gather} & 543125 & 0.95  &  	 \\
    \texttt{gather} & 543127 & 1	  &  	 \\
    \texttt{gather} & 543131 & 1	  &  	 \\
    \texttt{gather} & 543132 & 1	  &  	 \\
    \texttt{gather} & 543134 & 1	  &  	 \\
    \texttt{gather} & 543137 & 1	  &  Replaced (very difficult, no successful plays). 	 \\
    \texttt{jump} & 542276 & 0.97  &  	 \\
    \texttt{throw} & 542612 & 0.62	  & Replaced (due to bug, now fixed).	 \\
    \texttt{navigate} & 542971 & 0.85  &  \\
    \texttt{navigate} & 542973 & 0.9  &  Replaced (very difficult, no successful plays).  \\
    \texttt{navigate} & 542979 & 1 &  	 \\
    \texttt{navigate} & 542984 & 1 &  	 \\
    \texttt{survive} & 543077 & 1  &  	 \\
    \texttt{survive} & 543080 & 1  &  	 \\
    \texttt{survive} & 543086 & 1  &  	 \\
    \texttt{bridge} & 542527 & 1	  &  	 \\
    \texttt{carry} & 542936 & 1	  & Replaced (due to bug, now fixed).	 \\
    \bottomrule
  \end{tabular}
\end{table}

\subsection{Time Limit}

To limit the amount of time a participant could spend on a world, the following time limits were used:

\begin{itemize}
    \item 15 minutes for \texttt{navigate} and \texttt{find}
    \item 10 minutes for \texttt{survive}, \texttt{gather}, \texttt{stack}, \texttt{carry} and \texttt{explore}
    \item 5 minutes for all remaining tasks
\end{itemize}

When evaluating the agent, the maximum roll-out length was limited to be the same as human participants. See Appendix \ref{app:scoring} for more information on evaluation and scoring.

\section{Human evaluation}\label{app:humans}
\subsection{Selection and compensation}

Thirty participants were drawn from a pool of volunteers who indicated interest and an ability to commit 10 hours to the study during a week-long time-frame. Participants were asked to sign a participant consent form that included information regarding the purpose, procedure, risks and discomforts, potential benefits, costs, payment, confidentiality, and the subject’s rights during and after the study (see AdultConsentForm in the supplemental materials).

Risks and discomforts included cybersickness and short-term effects following VR use. The participants were also warned that there is a small chance that their identities could be inferred from their anonymized motion data. Participants were given an option to either be thanked or to remain entirely anonymous. All personally identifying information (i.e. names) has been removed from the dataset of human motion data (replaced with a random unique identifier).

Participants received payment in one of two forms:

\begin{enumerate}
    \item Any participant who did not already own an Oculus Quest 2 headset received a new one along with a supplemental battery pack, and any hours spent over the required 10 hours were reimbursed at \$30/hour.
    \item Any participant who already had an Oculus headset received a supplemental battery pack and was reimbursed at \$30/hour for the full 10 hours and any time they spent over that.
\end{enumerate}

The rate of \$30/hour was set to be fair to all participants regardless of whether they had an Oculus headset to start: an Oculus Quest 2 headset costs \$300, which is commensurate with a \$30/hour reimbursement over a 10-hour study period. Participants were also offered prizes of \$200, \$100, and \$50 for the top three performers to incentivize focus and maximum effort during the evaluation worlds, since the gameplay can become somewhat monotonous. These amounts were selected as a balance between incentivizing focus while avoiding causing any undue stress to participants. If a participant didn’t finish the required 10 hours, they were given the option to send back their headset (if applicable) and get reimbursed for the time spent or pay us back for the difference in hours. In total, approximately \$12,000 was spent on participant compensation between Oculus headsets, supplemental battery packs, and cash prizes/reimbursements. 

\subsection{Instruction and feedback}

For the first 3 hours of the study, participants were asked to set up their Oculus headsets by installing our Avalon APK. They received a document with setup instructions as well as information about basic game mechanics such as the game controls and the tools and animals they would encounter in the environment (also included in the supplemental materials, see ``Oculus Setup Practice Instructions''). Participants were then asked to complete at least 2 practice worlds for each of the 20 tasks. This data was not included in the evaluation set, and is not included in the recorded human data.

\begin{figure}
    \centering
    \includegraphics[width=1\linewidth]{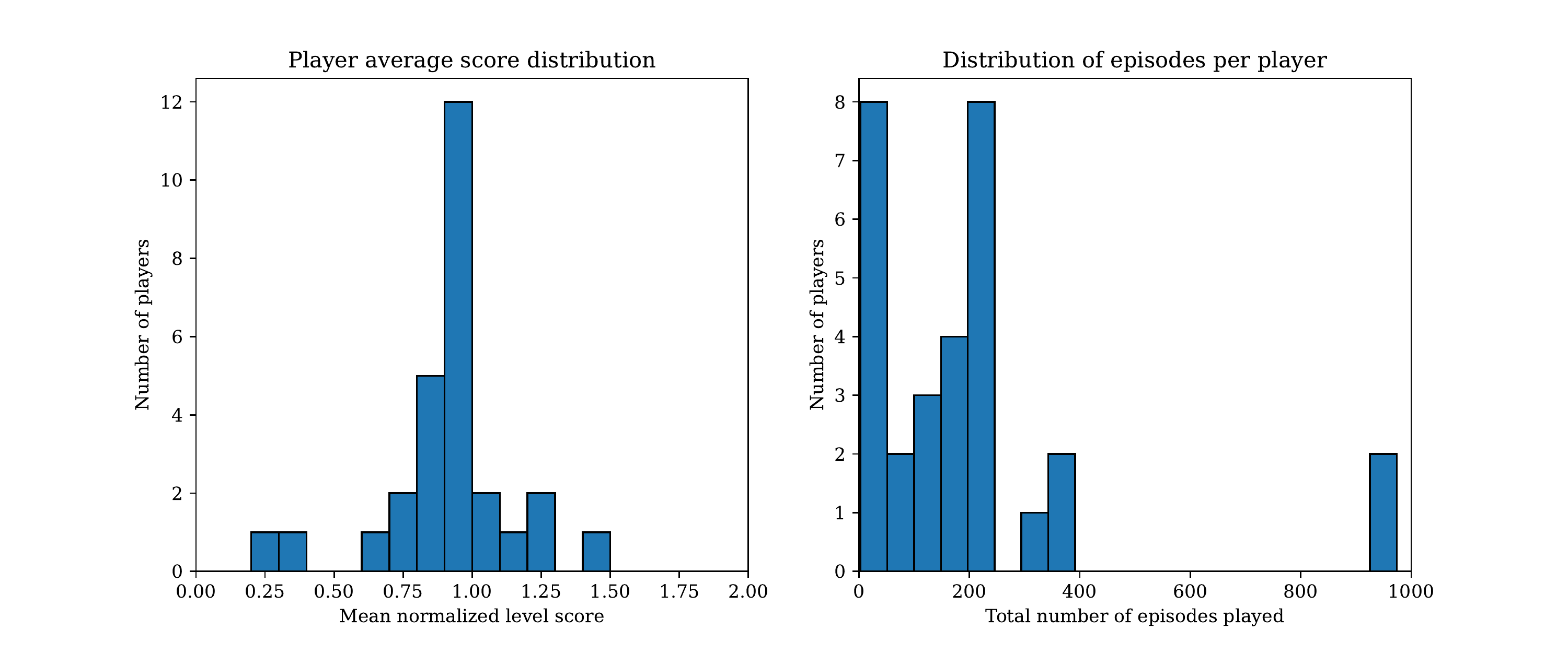}
    \caption{
    \textbf{Left}: Distribution of aggregated scores for each player. \textbf{Right}: Distribution of number of episodes played. Two users completed almost every evaluation level, while most completed around 200 levels.
    }
    \label{fig:human_score_eps_per_player}
\end{figure}

For the next 7 hours of the study, participants were asked to complete a series of evaluation worlds chosen at random from the pool of 1,000 evaluation worlds. No participant ever saw the same level twice. Two participants played almost 1,000 levels, and most players completed around 200 levels (Figure~\ref{fig:human_score_eps_per_player}), for a total of 6,145 played episodes. Participant motion data was captured during these evaluation worlds. Participants were allowed to ``reset'' if they ended up failing or getting stuck in a world. These resets were not counted towards the 5 playthroughs per world. This ability to reset was added to reduce stress on the human participants (whether due to perceived failure or real-life interruptions).

In a feedback survey after the study, participants reported an average overall satisfaction with the study of 4.57 out of 5 and 30 out of 32 of the participants said they would like to be considered for future studies.

\subsection{Analysis of performance}

\begin{figure}
    \centering
    \includegraphics[width=0.7\linewidth]{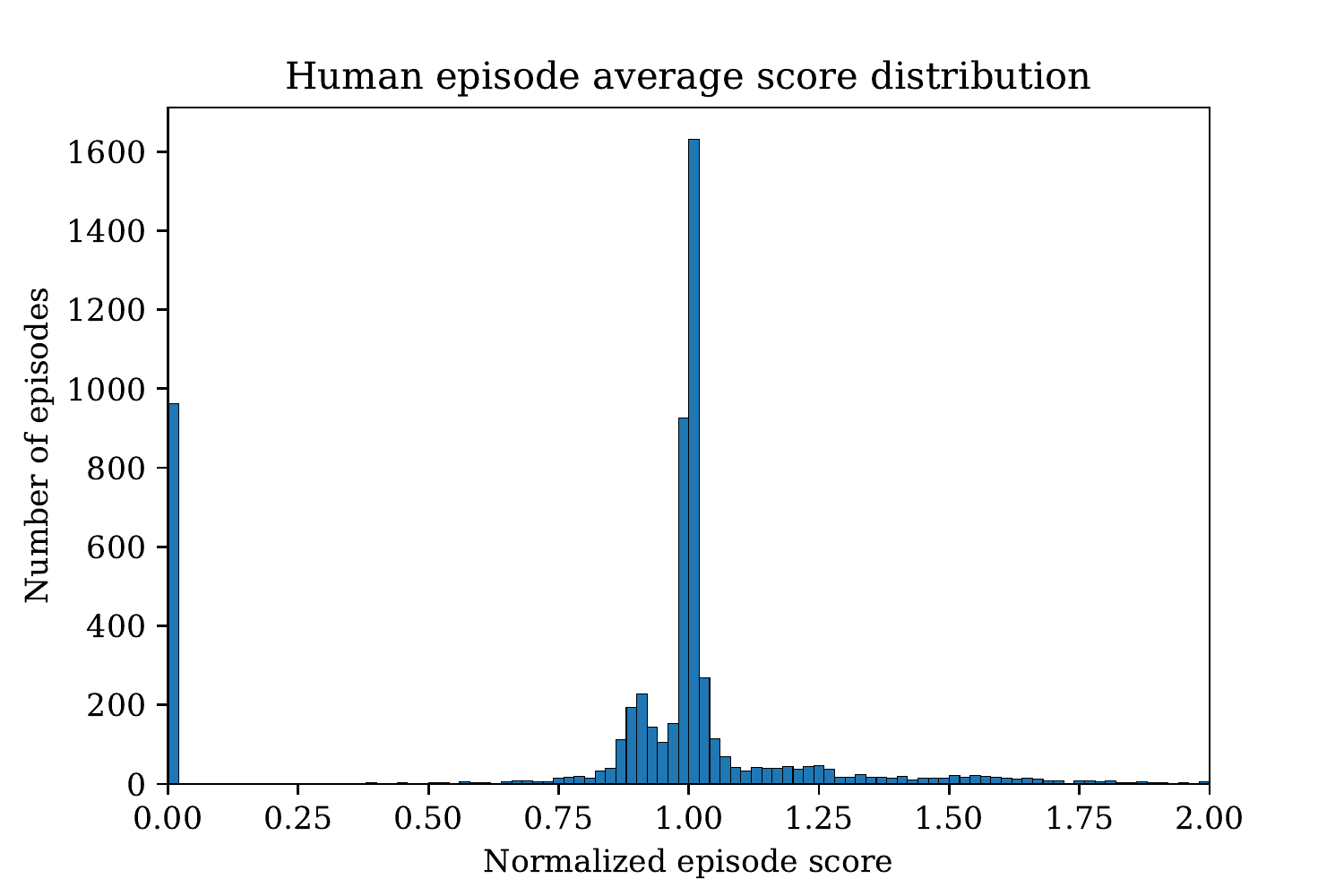}
    \caption{
    Distribution of scores for each episode. Most players were able to eat all the available food in about the same time for most episodes, leading to a narrow distribution of scores.
    }
    \label{fig:human_score_per_episode}
\end{figure}

Most levels were quite simple, leading to final episode scores clustered tightly around one (Figure~\ref{fig:human_score_per_episode}). Aggregated player performance was quite similar as well (Figure~\ref{fig:human_score_eps_per_player}), with low performing users playing on just a few high difficulty levels.

\begin{figure}
    \centering
    \includegraphics[width=0.7\linewidth]{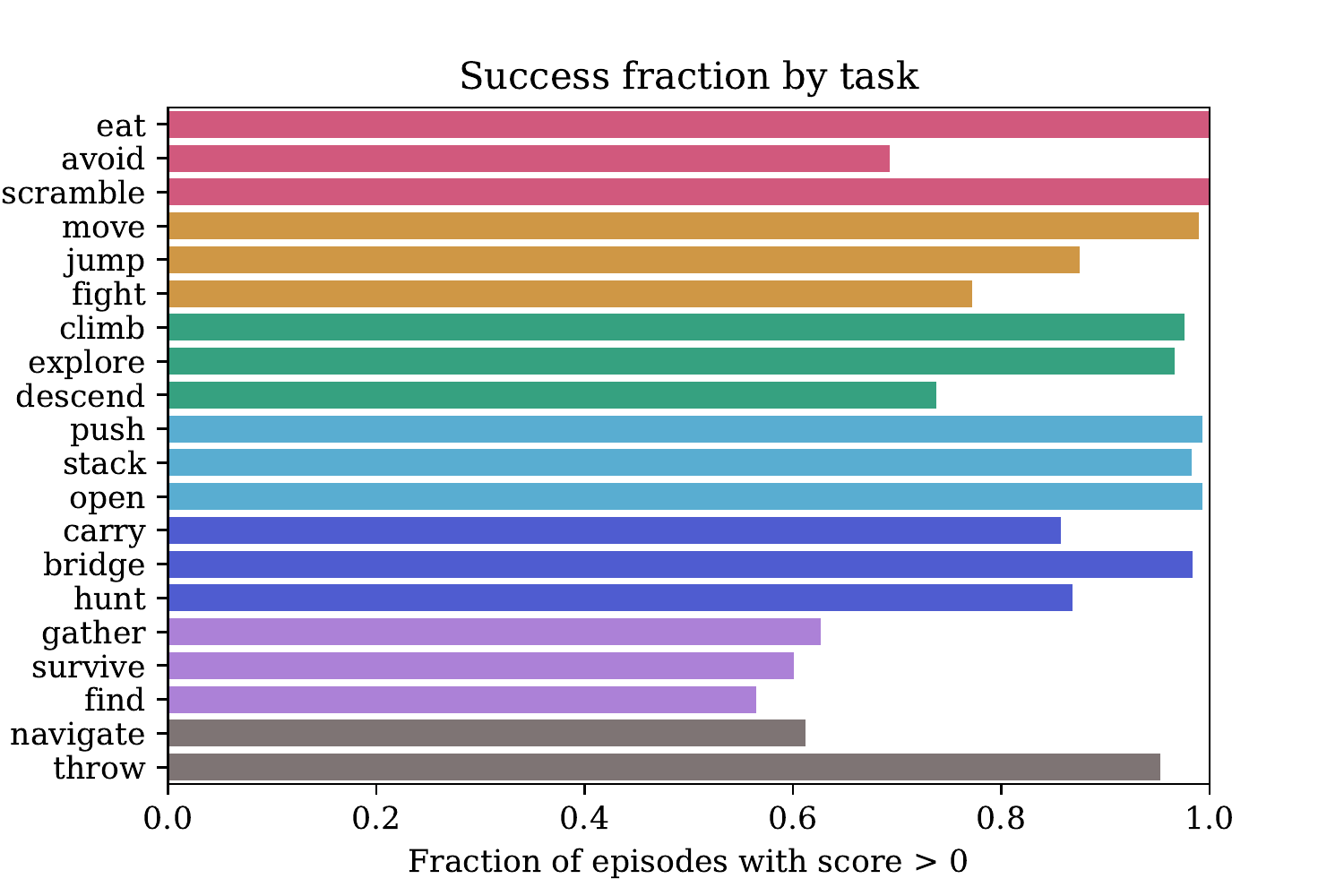}
    \caption{
    Success rate of players on each task. Success is defined as completing an episode with a score greater than 0.
    }
    \label{fig:human_success_fraction}
\end{figure}

In 962 episodes, the player either died during the episode or reached the time limit without eating food, getting a final score of zero. The remaining 5,183 episodes had a score greater than zero, a condition we refer to as a ``success.'' The distribution of these successes across tasks is shown in Figure~\ref{fig:human_success_fraction}. Human performance is fairly consistent across most tasks, with some of the compositional tasks presenting more of a challenge.

\section{Curriculum}\label{app:curriculum}

All task generators are parameterized, at the highest level, by the task id and the “difficulty” $d_t$ (a float in the range [0, 1]). Each environment generation worker process maintains a mapping from task id to the current maximum difficulty $d_t$ (all initialized at 0), and generates a level from a uniform distribution $U[0,d_t]$. When an agent succeeds (or fails) at a generated environment, the $d_t$ is increased (or decreased) by  $H_t$, a hyper-parameter that controls how quickly the task curriculum adjusts. An agent succeeds at its environment if it eats all of the available food before the episode timer is finished.

\section{Hyperparameters}\label{app:hyperparams}
See table \ref{tab:hyperparameters} for hyperparameters used for training PPO, IMPALA, and Dreamer.

\begin{table}[h!]
  \caption{Hyperparameters used for training.}
  \label{tab:hyperparameters}
  \centering
\begin{tabular}{lccc}
\toprule
\textbf{Parameter} & \textbf{PPO} & \textbf{IMPALA} & \textbf{Dreamer} \\
\midrule
Optimizer & Adam & RMSProp & Adam \\
\multirow{3}{*}{Learning rate} & \multirow{3}{*}{2.5e-4} & \multirow{3}{*}{1.5e-4} & model=1e-4 \\
& & & value=1e-4 \\
& & & actor=1e-5  \\
\multirow{2}{*}{Other optimizer params} & \multirow{2}{*}{$\epsilon=$1e-5} & $\alpha=0.95$ & $\epsilon=$1e-5 \\
 & & $\epsilon=0.005$ & weight decay=1e-6 \\
Grad norm clipping & 0.5 & 200 & 100 \\
Baseline cost & 1 & 2 & 1 \\
Entropy cost & 1.5e-4 & 5e-3 & 2e-3 \\
Discount factor & 0.99 & 0.998 & .99 \\
GAE Lambda & 0.83 & - & .95 \\
IS clip range & 0.03 & - & - \\
Batch size & 16 & 32 & 116 \\
Unroll length & 200 & 100 & 30 \\
Task difficulty update ($H_t$) & 3e-4 & 5e-3 & 6e-4 \\
Movement penalty coefficient ($C_T$) & 1e-8 & 1e-8 & 1e-8 \\
Frame penalty ($P_{\rm frame}$) & 1e-4 & 1e-4 & 1e-4 \\
Actor gradients & - & - & REINFORCE \\
\bottomrule
\end{tabular}
\end{table}

\section{Training compute}\label{app:compute}
Our training took place in containers, with 8 containers to a machine. One GPU was assigned per container. The whole machine provides 8x Nvidia GeForce RTX 3090 (with 24GB RAM), and is powered by 2x AMD EPYC 7313 CPUs (with 16 cores/32 threads each) with 256GB RAM total. Training time was approximately 30 hours for a single run.

\section{Results for other training runs}\label{app:other_results}
Table~\ref{tab:optimality_gap_scores} shows the optimality gap scores corresponding to our main experimental runs (whereas Table~\ref{tab:scores} reports average scores).
The same trends are apparent in both tables: the compositional tasks are much more difficult than most of the basic tasks, and the gap between human scores and even the best-performing networks are emphasized here.

\begin{table}
    \centering
\begin{tabular}{llllll}
    \toprule
    \textbf{Task} & \textbf{PPO} & \textbf{Dreamer} & \multicolumn{3}{c}{\textbf{IMPALA}} \\
    
    \cmidrule{4-6}
        &   50m steps  &   50m steps &  \multicolumn{1}{c}{50m steps}  & \multicolumn{1}{c}{500m steps}   & \multicolumn{1}{c}{50m steps}\\
        &   With curr. &   With curr.  &  \multicolumn{1}{c}{With curr.}  & \multicolumn{1}{c}{With curr.}   & \multicolumn{1}{c}{No curr.}   \\
\midrule
 \texttt{eat}       & 0.402 $\pm$ 0.067 & 0.445 $\pm$ 0.065  & 0.357 $\pm$ 0.062 & 0.375 $\pm$ 0.097        & 0.999 $\pm$ 0.001                 \\
 \texttt{move}      & 0.741 $\pm$ 0.062 & 0.721 $\pm$ 0.071  & 0.640 $\pm$ 0.062 & 0.603 $\pm$ 0.076        & 1.000 $\pm$ 0.000                 \\
 \texttt{jump}      & 0.798 $\pm$ 0.050 & 0.806 $\pm$ 0.058  & 0.709 $\pm$ 0.056 & 0.725 $\pm$ 0.072        & 1.000 $\pm$ 0.000                 \\
 \texttt{climb}     & 0.816 $\pm$ 0.043 & 0.809 $\pm$ 0.051  & 0.782 $\pm$ 0.049 & 0.675 $\pm$ 0.074        & 1.000 $\pm$ 0.000                 \\
 \texttt{descend}   & 0.834 $\pm$ 0.043 & 0.751 $\pm$ 0.058  & 0.843 $\pm$ 0.044 & 0.784 $\pm$ 0.059        & 1.000 $\pm$ 0.000                 \\
 \texttt{scramble}  & 0.719 $\pm$ 0.054 & 0.627 $\pm$ 0.058  & 0.573 $\pm$ 0.062 & 0.444 $\pm$ 0.070        & 1.000 $\pm$ 0.000                 \\
 \texttt{stack}     & 0.921 $\pm$ 0.036 & 0.890 $\pm$ 0.043  & 0.874 $\pm$ 0.036 & 0.891 $\pm$ 0.055        & 1.000 $\pm$ 0.000                 \\
 \texttt{bridge}    & 0.956 $\pm$ 0.027 & 0.907 $\pm$ 0.045  & 0.924 $\pm$ 0.029 & 0.912 $\pm$ 0.049        & 1.000 $\pm$ 0.000                 \\
 \texttt{push}      & 0.899 $\pm$ 0.039 & 0.874 $\pm$ 0.053  & 0.867 $\pm$ 0.043 & 0.872 $\pm$ 0.054        & 1.000 $\pm$ 0.000                 \\
 \texttt{throw}     & 1.000 $\pm$ 0.000 & 1.000 $\pm$ 0.000  & 1.000 $\pm$ 0.000 & 1.000 $\pm$ 0.000        & 1.000 $\pm$ 0.000                 \\
 \texttt{hunt}      & 0.961 $\pm$ 0.024 & 0.942 $\pm$ 0.028  & 0.933 $\pm$ 0.029 & 0.880 $\pm$ 0.051        & 1.000 $\pm$ 0.000                 \\
 \texttt{fight}     & 0.831 $\pm$ 0.052 & 0.746 $\pm$ 0.076  & 0.789 $\pm$ 0.052 & 0.724 $\pm$ 0.075        & 1.000 $\pm$ 0.000                 \\
 \texttt{avoid}     & 0.697 $\pm$ 0.170 & 0.634 $\pm$ 0.118  & 0.551 $\pm$ 0.159 & 0.510 $\pm$ 0.102        & 1.000 $\pm$ 0.000                 \\
 \texttt{explore}   & 0.827 $\pm$ 0.047 & 0.831 $\pm$ 0.048  & 0.803 $\pm$ 0.048 & 0.760 $\pm$ 0.069        & 1.000 $\pm$ 0.000                 \\
 \texttt{open}      & 0.948 $\pm$ 0.024 & 0.892 $\pm$ 0.041  & 0.907 $\pm$ 0.034 & 0.899 $\pm$ 0.046        & 1.000 $\pm$ 0.000                 \\
 \texttt{carry}     & 0.932 $\pm$ 0.031 & 0.940 $\pm$ 0.028  & 0.913 $\pm$ 0.032 & 0.891 $\pm$ 0.057        & 1.000 $\pm$ 0.000                 \\
\midrule
 \texttt{navigate}  & 1.000 $\pm$ 0.000 & 1.000 $\pm$ 0.000  & 0.988 $\pm$ 0.010 & 0.963 $\pm$ 0.032        & 1.000 $\pm$ 0.000                 \\
 \texttt{find}      & 0.998 $\pm$ 0.003 & 1.000 $\pm$ 0.000  & 0.987 $\pm$ 0.014 & 0.987 $\pm$ 0.017        & 1.000 $\pm$ 0.000                 \\
 \texttt{survive}   & 0.957 $\pm$ 0.013 & 0.956 $\pm$ 0.014  & 0.950 $\pm$ 0.015 & 0.915 $\pm$ 0.028        & 1.000 $\pm$ 0.000                 \\
 \texttt{gather}    & 0.979 $\pm$ 0.010 & 0.980 $\pm$ 0.012  & 0.970 $\pm$ 0.010 & 0.968 $\pm$ 0.014        & 1.000 $\pm$ 0.000                 \\
\midrule
 all basic & 0.830 $\pm$ 0.017 & 0.801 $\pm$ 0.016  & 0.779 $\pm$ 0.016 & 0.747 $\pm$ 0.019        & 1.000 $\pm$ 0.000                 \\
 all comp. & 0.983 $\pm$ 0.004 & 0.984 $\pm$ 0.005  & 0.974 $\pm$ 0.007 & 0.958 $\pm$ 0.013        & 1.000 $\pm$ 0.000                 \\
 all       & 0.861 $\pm$ 0.014 & 0.838 $\pm$ 0.012  & 0.818 $\pm$ 0.013 & 0.789 $\pm$ 0.015        & 1.000 $\pm$ 0.000                 \\
\bottomrule
\end{tabular}
    \caption{Optimality gap results for Table~\ref{tab:scores}. Lower scores are better.}
    \label{tab:optimality_gap_scores}
\end{table}

Tables \ref{tab:taskwise_agent_scores} and \ref{tab:taskwise_optimality_gap_scores} show the average scores and optimality gap scores respectively for IMPALA under each of the four training conditions, averaged over three training runs with different seeds. A few salient conclusions may be drawn from this data. First, comparing MT-TB with the single task baselines ST-B, one can see that the similarity of environments has enabled significant transfer learning. Several tasks like \texttt{jump} and \texttt{climb} are too challenging to learn on their own, but may be achieved by training on other tasks. While the performance of basic tasks such \texttt{eat} and \texttt{move} somewhat lower in the multi-task setting than the single task setting, the difference is small enough that it is more likely due to fewer exposures to each task in the multi task case, rather than being a case of catastrophic forgetting. 

Second, we note that training on only the basic tasks seems to be the most effective use of training time. While an agent training on all tasks or even just the compositional tasks does learn a broad variety of basic skills (for example, despite never seeing these levels, MT-TC has significant performance on \texttt{eat} and \texttt{open}), these agents do not outperform an agent trained on just the basic tasks when measured on either the basic or compositional aggregate scores. We suspect that these tasks may be too difficult for the agent to make much progress at this stage in training.

\begin{table}
    \centering
\begin{tabular}{lllll}
\toprule
 \textbf{Task}      & \textbf{MT-TB}    & \textbf{MT-TA}    & \textbf{MT-TC}    & \textbf{ST-B}     \\
\midrule
 \texttt{eat}       & 0.716 $\pm$ 0.062 & 0.657 $\pm$ 0.070 & 0.236 $\pm$ 0.051 & 0.751 $\pm$ 0.060 \\
 \texttt{move}      & 0.399 $\pm$ 0.062 & 0.370 $\pm$ 0.064 & 0.018 $\pm$ 0.014 & 0.397 $\pm$ 0.059 \\
 \texttt{jump}      & 0.309 $\pm$ 0.056 & 0.288 $\pm$ 0.056 & 0.008 $\pm$ 0.011 & 0.112 $\pm$ 0.033 \\
 \texttt{climb}     & 0.229 $\pm$ 0.047 & 0.144 $\pm$ 0.043 & 0.000 $\pm$ 0.000 & 0.000 $\pm$ 0.000 \\
 \texttt{descend}   & 0.173 $\pm$ 0.045 & 0.128 $\pm$ 0.037 & 0.002 $\pm$ 0.003 & 0.000 $\pm$ 0.000 \\
 \texttt{scramble}  & 0.467 $\pm$ 0.064 & 0.252 $\pm$ 0.053 & 0.006 $\pm$ 0.008 & 0.630 $\pm$ 0.050 \\
 \texttt{stack}     & 0.130 $\pm$ 0.038 & 0.115 $\pm$ 0.037 & 0.000 $\pm$ 0.000 & 0.000 $\pm$ 0.000 \\
 \texttt{bridge}    & 0.076 $\pm$ 0.028 & 0.044 $\pm$ 0.026 & 0.000 $\pm$ 0.000 & 0.000 $\pm$ 0.000 \\
 \texttt{push}      & 0.150 $\pm$ 0.045 & 0.092 $\pm$ 0.035 & 0.000 $\pm$ 0.000 & 0.000 $\pm$ 0.000 \\
 \texttt{throw}     & 0.000 $\pm$ 0.000 & 0.000 $\pm$ 0.000 & 0.000 $\pm$ 0.000 & 0.000 $\pm$ 0.000 \\
 \texttt{hunt}      & 0.071 $\pm$ 0.029 & 0.030 $\pm$ 0.020 & 0.003 $\pm$ 0.005 & 0.250 $\pm$ 0.051 \\
 \texttt{fight}     & 0.235 $\pm$ 0.053 & 0.212 $\pm$ 0.050 & 0.002 $\pm$ 0.004 & 0.000 $\pm$ 0.000 \\
 \texttt{avoid}     & 0.603 $\pm$ 0.144 & 0.326 $\pm$ 0.055 & 0.000 $\pm$ 0.000 & 0.000 $\pm$ 0.000 \\
 \texttt{explore}   & 0.213 $\pm$ 0.050 & 0.190 $\pm$ 0.044 & 0.044 $\pm$ 0.025 & 0.247 $\pm$ 0.054 \\
 \texttt{open}      & 0.097 $\pm$ 0.034 & 0.063 $\pm$ 0.025 & 0.009 $\pm$ 0.010 & 0.083 $\pm$ 0.029 \\
 \texttt{carry}     & 0.089 $\pm$ 0.033 & 0.090 $\pm$ 0.037 & 0.005 $\pm$ 0.006 & 0.000 $\pm$ 0.000 \\
\midrule
 \texttt{navigate}  & 0.012 $\pm$ 0.011 & 0.000 $\pm$ 0.000 & 0.000 $\pm$ 0.000 & 0.000 $\pm$ 0.000 \\
 \texttt{find}      & 0.015 $\pm$ 0.015 & 0.008 $\pm$ 0.011 & 0.000 $\pm$ 0.000 & 0.000 $\pm$ 0.000 \\
 \texttt{survive}   & 0.050 $\pm$ 0.014 & 0.037 $\pm$ 0.011 & 0.010 $\pm$ 0.005 & 0.031 $\pm$ 0.013 \\
 \texttt{gather}    & 0.030 $\pm$ 0.010 & 0.019 $\pm$ 0.008 & 0.000 $\pm$ 0.000 & 0.000 $\pm$ 0.000 \\
\midrule
 all basic & 0.247 $\pm$ 0.017 & 0.188 $\pm$ 0.012 & 0.021 $\pm$ 0.004 & -                 \\
 all comp. & 0.027 $\pm$ 0.006 & 0.016 $\pm$ 0.005 & 0.002 $\pm$ 0.001 & -                 \\
 all       & 0.203 $\pm$ 0.012 & 0.153 $\pm$ 0.010 & 0.017 $\pm$ 0.003 & -                 \\
\bottomrule
\end{tabular}
    \caption{
    Mean-human normalized performance for RL agents trained with protocols defined in Section \ref{sec:training-protocols}. MT-TB is trained on the 16 simple tasks, MT-TA is trained on all 20 tasks, and MT-TC is trained on the four compositional tasks. All agents are trained for 50M steps using IMPALA. ST-B results are trained only on a single task for 50M steps, and are evaluated only on that same task.
    }
    \label{tab:taskwise_agent_scores}
\end{table}

\begin{table}
    \centering
\begin{tabular}{lllll}
\toprule
 \textbf{Task}      & \textbf{MT-TB}    & \textbf{MT-TA}    & \textbf{MT-TC}    & \textbf{ST-B}     \\
\midrule
 \texttt{eat}       & 0.357 $\pm$ 0.062 & 0.451 $\pm$ 0.070 & 0.787 $\pm$ 0.051 & 0.316 $\pm$ 0.060 \\
 \texttt{move}      & 0.640 $\pm$ 0.062 & 0.673 $\pm$ 0.064 & 0.982 $\pm$ 0.014 & 0.616 $\pm$ 0.059 \\
 \texttt{jump}      & 0.709 $\pm$ 0.056 & 0.736 $\pm$ 0.056 & 0.994 $\pm$ 0.011 & 0.889 $\pm$ 0.033 \\
 \texttt{climb}     & 0.782 $\pm$ 0.047 & 0.867 $\pm$ 0.043 & 1.000 $\pm$ 0.000 & 1.000 $\pm$ 0.000 \\
 \texttt{descend}   & 0.843 $\pm$ 0.045 & 0.883 $\pm$ 0.037 & 0.998 $\pm$ 0.003 & 1.000 $\pm$ 0.000 \\
 \texttt{scramble}  & 0.573 $\pm$ 0.064 & 0.767 $\pm$ 0.053 & 0.994 $\pm$ 0.008 & 0.386 $\pm$ 0.050 \\
 \texttt{stack}     & 0.874 $\pm$ 0.038 & 0.896 $\pm$ 0.037 & 1.000 $\pm$ 0.000 & 1.000 $\pm$ 0.000 \\
 \texttt{bridge}    & 0.924 $\pm$ 0.028 & 0.961 $\pm$ 0.026 & 1.000 $\pm$ 0.000 & 1.000 $\pm$ 0.000 \\
 \texttt{push}      & 0.867 $\pm$ 0.045 & 0.915 $\pm$ 0.035 & 1.000 $\pm$ 0.000 & 1.000 $\pm$ 0.000 \\
 \texttt{throw}     & 1.000 $\pm$ 0.000 & 1.000 $\pm$ 0.000 & 1.000 $\pm$ 0.000 & 1.000 $\pm$ 0.000 \\
 \texttt{hunt}      & 0.933 $\pm$ 0.029 & 0.974 $\pm$ 0.020 & 0.997 $\pm$ 0.005 & 0.767 $\pm$ 0.051 \\
 \texttt{fight}     & 0.789 $\pm$ 0.053 & 0.812 $\pm$ 0.050 & 0.998 $\pm$ 0.004 & 1.000 $\pm$ 0.000 \\
 \texttt{avoid}     & 0.551 $\pm$ 0.144 & 0.704 $\pm$ 0.055 & 1.000 $\pm$ 0.000 & 1.000 $\pm$ 0.000 \\
 \texttt{explore}   & 0.803 $\pm$ 0.050 & 0.822 $\pm$ 0.044 & 0.961 $\pm$ 0.025 & 0.773 $\pm$ 0.054 \\
 \texttt{open}      & 0.907 $\pm$ 0.034 & 0.938 $\pm$ 0.025 & 0.991 $\pm$ 0.010 & 0.918 $\pm$ 0.029 \\
 \texttt{carry}     & 0.913 $\pm$ 0.033 & 0.921 $\pm$ 0.037 & 0.995 $\pm$ 0.006 & 1.000 $\pm$ 0.000 \\
\midrule
 \texttt{navigate}  & 0.988 $\pm$ 0.011 & 1.000 $\pm$ 0.000 & 1.000 $\pm$ 0.000 & 1.000 $\pm$ 0.000 \\
 \texttt{find}      & 0.987 $\pm$ 0.015 & 0.994 $\pm$ 0.011 & 1.000 $\pm$ 0.000 & 1.000 $\pm$ 0.000 \\
 \texttt{survive}   & 0.950 $\pm$ 0.014 & 0.963 $\pm$ 0.011 & 0.990 $\pm$ 0.005 & 0.969 $\pm$ 0.013 \\
 \texttt{gather}    & 0.970 $\pm$ 0.010 & 0.981 $\pm$ 0.008 & 1.000 $\pm$ 0.000 & 1.000 $\pm$ 0.000 \\
\midrule
 all basic & 0.779 $\pm$ 0.017 & 0.833 $\pm$ 0.012 & 0.981 $\pm$ 0.004 & -                 \\
 all comp. & 0.974 $\pm$ 0.006 & 0.984 $\pm$ 0.005 & 0.998 $\pm$ 0.001 & -                 \\
 all       & 0.818 $\pm$ 0.012 & 0.863 $\pm$ 0.010 & 0.984 $\pm$ 0.003 & -                 \\
\bottomrule
\end{tabular}
    \caption{
    Optimality gap results for Table \ref{tab:taskwise_agent_scores}. Lower scores are better.
    }
    \label{tab:taskwise_optimality_gap_scores}
\end{table}

\begin{figure}
    \centering
    \includegraphics[width=1.0\linewidth]{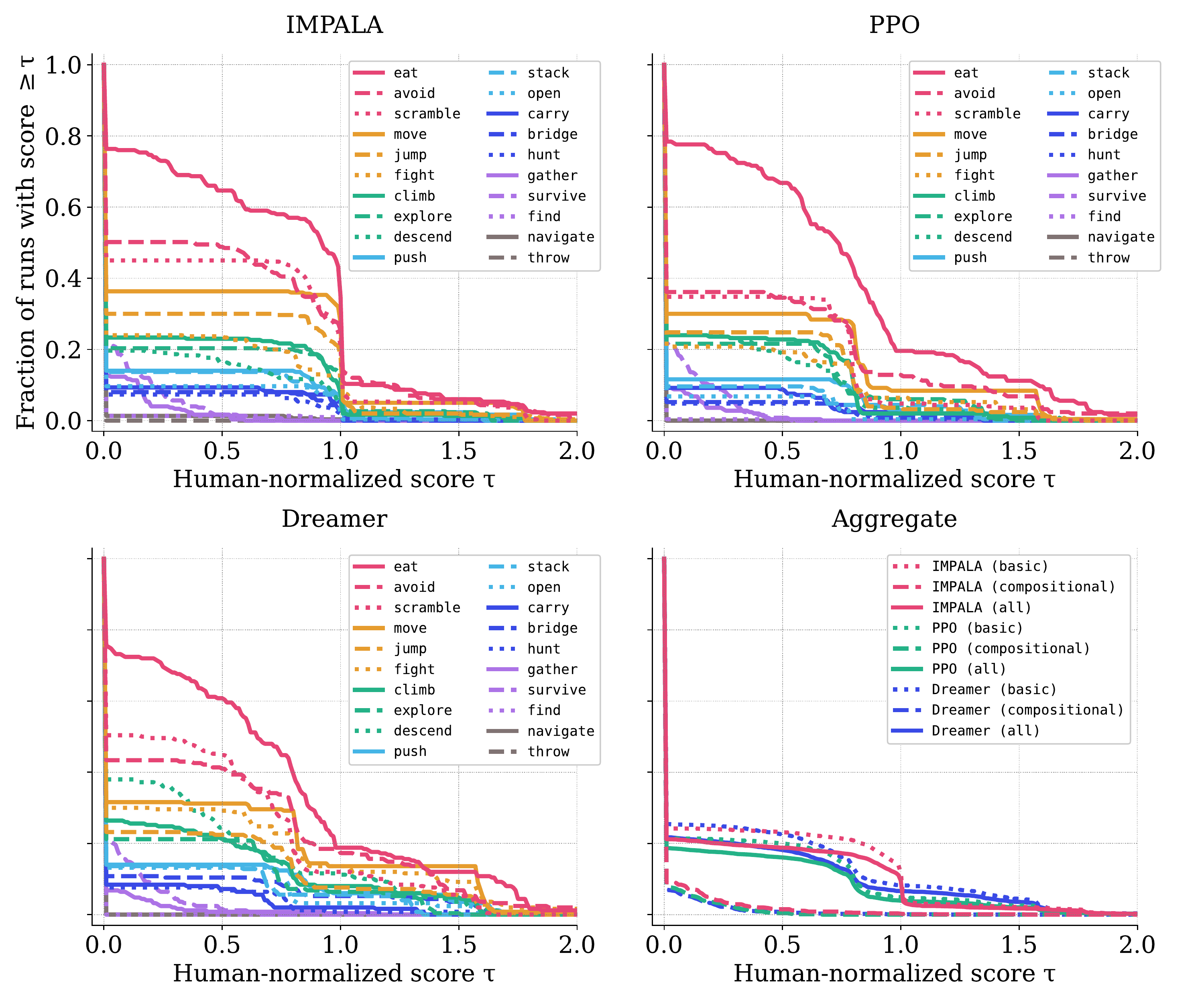}
    \caption{
    Results for RL agents trained on all 16 basic tasks and evaluated on each of Avalon's 20 tasks (MT-TB).
    Scores shown are average scores, normalized against mean human performance.
    Top row and bottom left: scores for IMPALA, PPO and Dreamer on each task.
    Bottom right: aggregate scores averaged over the 16 basic tasks, the four compositional tasks, and all 20 tasks.
    }
    \label{fig:taskwise_agent_scores}
\end{figure}

In Figure \ref{fig:taskwise_agent_scores}, we show a performance profile by task for each network (IMPALA and PPO) on the MT-TB training setting (for which results are shown in Table \ref{tab:scores} in the main paper). For many tasks, humans had quite consistent performance, due to the simplicity of the tasks and sparse reward. This leads to the relatively flat horizontal sections of the graph extending from 0 to 1, where the y-axis indicates the fraction of these levels in which the agent got the single available food. For many tasks the scores drop off as they approach $\tau = 1$, as the tasks get harder and there is some more nuance to performance beyond the binary "found food or not" (e.g., from tasks that take longer and or where there is some risk to humans of dying). From the tasks with any amount of super-human performance, these seem likely to be cases where the episode was extremely short and the agent moved faster and oriented itself more quickly than the average human.

From the graphs in Figure \ref{fig:taskwise_agent_scores}, we can also see that most of the advantage for IMPALA over PPO comes from better performance on the basic tasks. Both networks perform quite poorly on the compositional tasks, mostly only succeeding at \texttt{survive}, which often has plentiful food, making nonzero scores easier to achieve. One can get a sense for the relative difficulties of the tasks. Some, like \texttt{open} and \texttt{avoid}, seem surprisingly easy, though this is likely due to relatively forgiving levels at low difficulties rather than due to highly effective agents.

\begin{figure}[h!]
    \centering
    \includegraphics[width=1\linewidth]{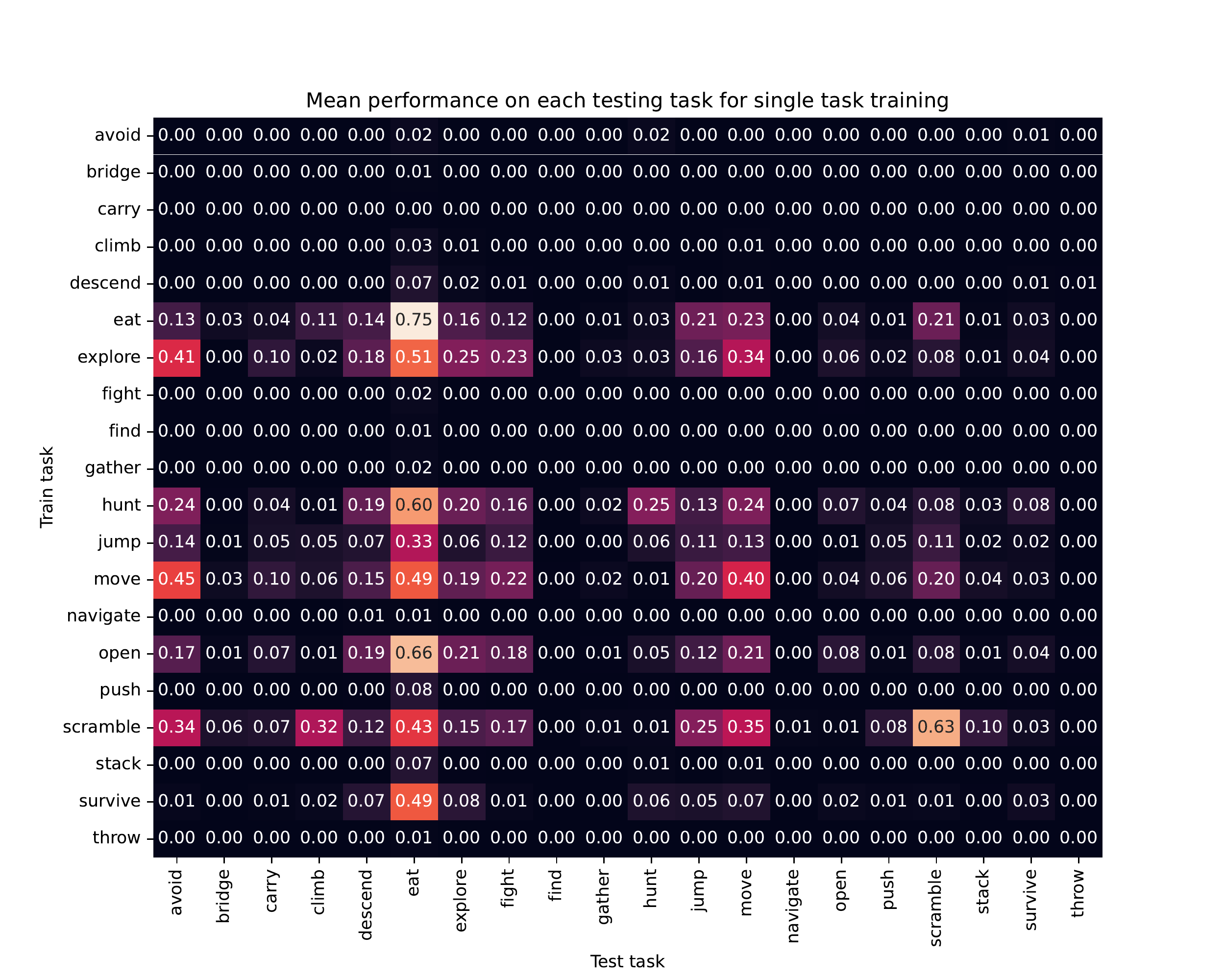}
    \caption{
    Mean human-normalized performance for IMPALA trained for 50M environment steps on a single task and tested on another task.
    }
    \label{fig:single_task_training_matrix}
\end{figure}

We visualize the generalization performance of single task training in Figure~\ref{fig:single_task_training_matrix}. One can see that training on other tasks readily transfers to \texttt{eat} or \texttt{move}, but that the agents are not able to make much progress on either the trained task or the other tasks.

\begin{figure}
    \centering
    \includegraphics[width=0.7\linewidth]{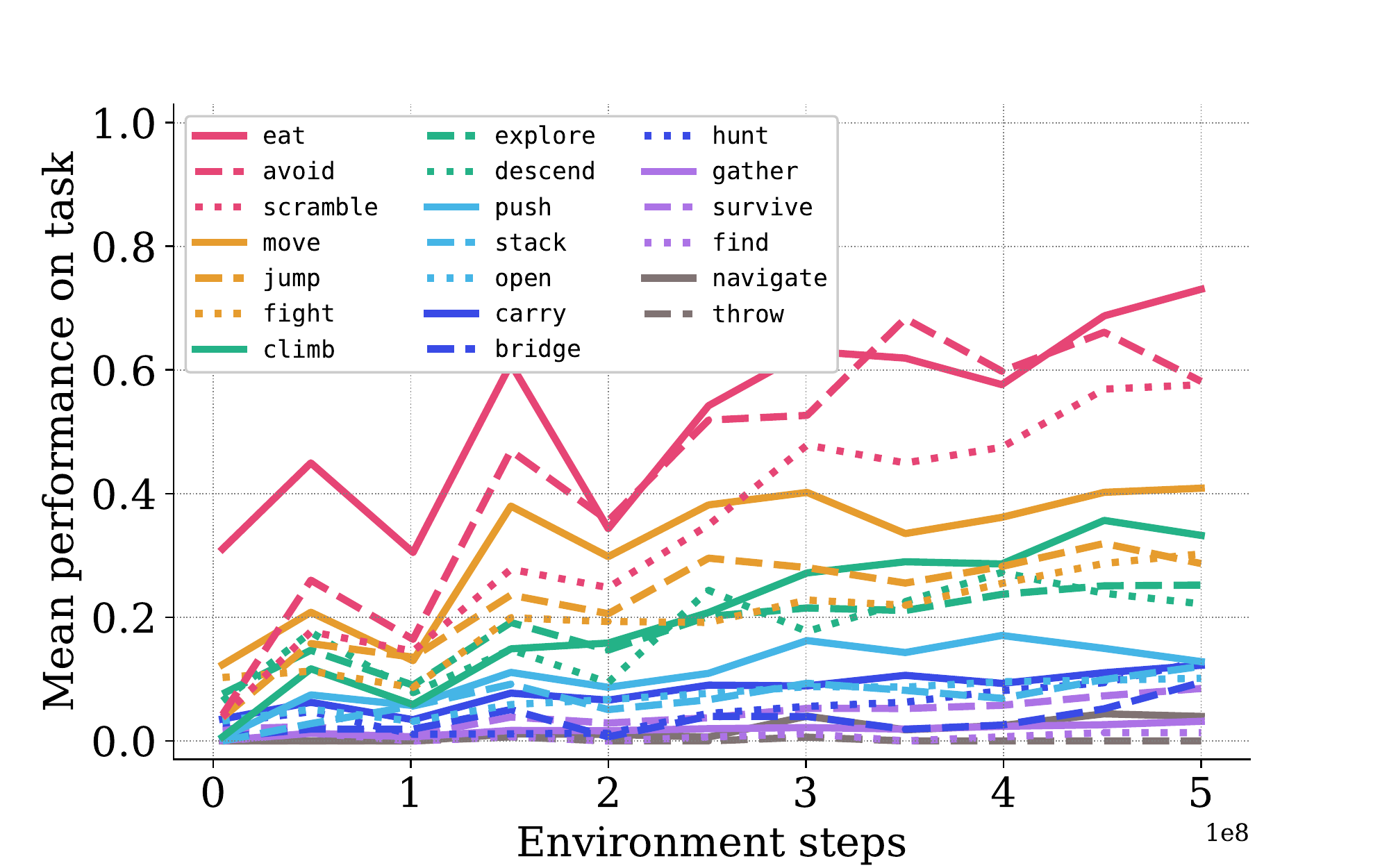}
    \caption{
    Learning curves for mean human normalized performance on each task, using IMPALA and MT-TB for 500m training runs with the curriculum. Each point is the average over 3 different training runs with different seeds.
    }
    \label{fig:learning_curve}
\end{figure}

We finally show learning curves for IMPALA using MT-TB in Figure \ref{fig:learning_curve}. One can see that performance on some tasks continues to improve throughout training, and so the agent may continue to benefit from more steps in the environment.

\section{Scoring}\label{app:scoring}
For both human players and RL agents, the score of each run is given by

\begin{equation}
    S = \max \left( 0, E_f - E_0  + T \times P_{\rm frame}  \cdot \right),
\end{equation}

where $E_0$ is the starting energy level, $E_f$ is the final energy level (when the last food is consumed, the agent dies or the episode times out), $T$ is the total number of episode frames remaining until timeout, and $P_{\rm frame}$ is a constant penalty per frame. For effective comparison with humans, we do not impose energy penalties for movement during evaluation (i.e. the $C_{\rm move}$ of Appendix \ref{app:reward} is zero), opting instead for the constant per-frame energy penalty. We use the same $P_{\rm frame}$ during training of 1e-4 when scoring runs and for each task, $E_0$ is always set to 1.0.

For most tasks, success is simply whether or not the player ate all of the food in the level (and there is usually just a single piece). For tasks which always have multiple pieces of food, namely \texttt{gather} and \texttt{survive}, a run is considered successful if the player ate all food, or ate at least one food and ended the task with more health than they started with.

All reported scores are normalized post hoc such that the average human performance on a given task is 1.0 and the performance of a random agent is 0.0.






\section{Simulator Performance}\label{app:performance}
\begin{figure}
     \centering
     \begin{subfigure}[b]{0.24\textwidth}
         \centering
         \includegraphics[width=\textwidth]{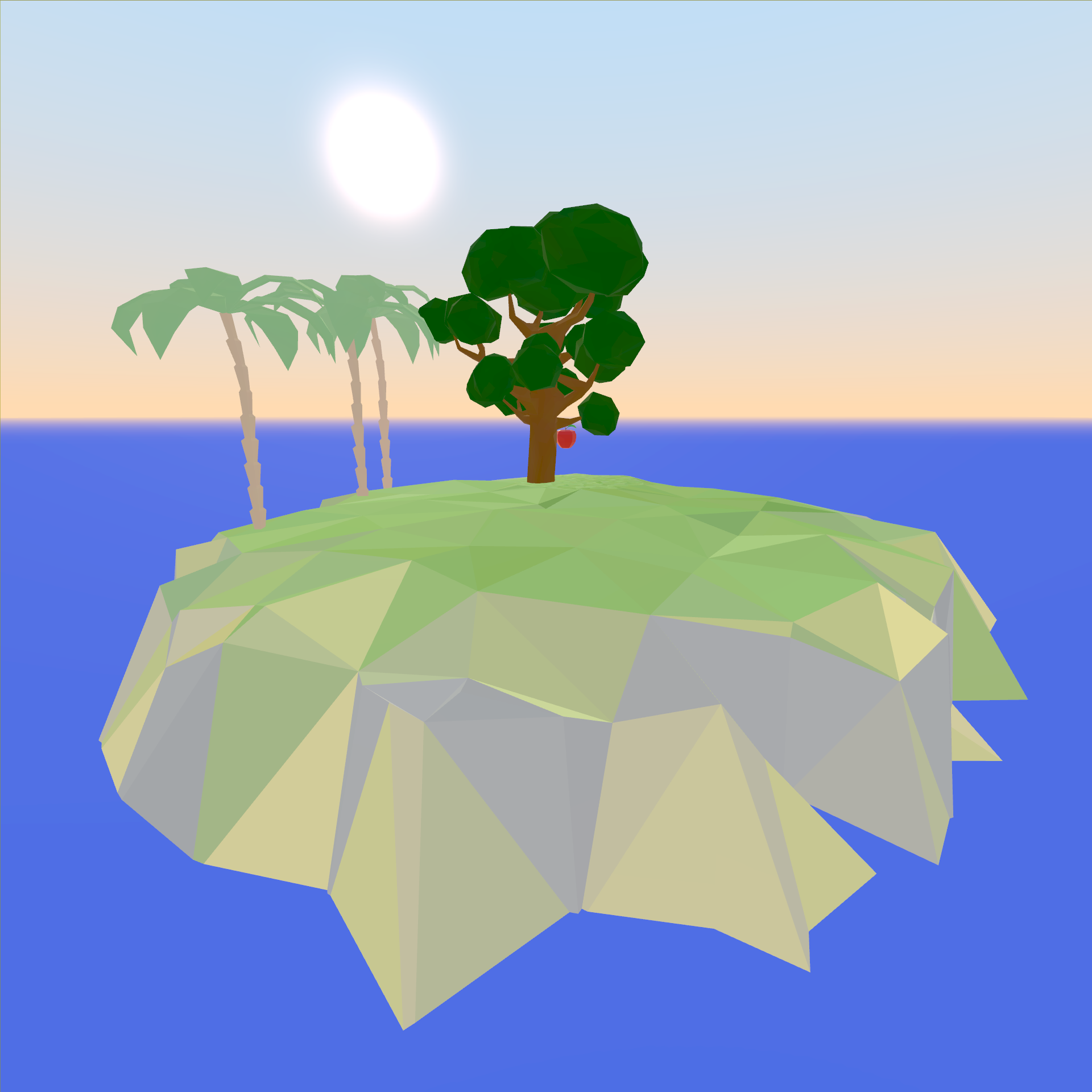}
     \end{subfigure}
     \hfill
     \begin{subfigure}[b]{0.24\textwidth}
         \centering
         \includegraphics[width=\textwidth]{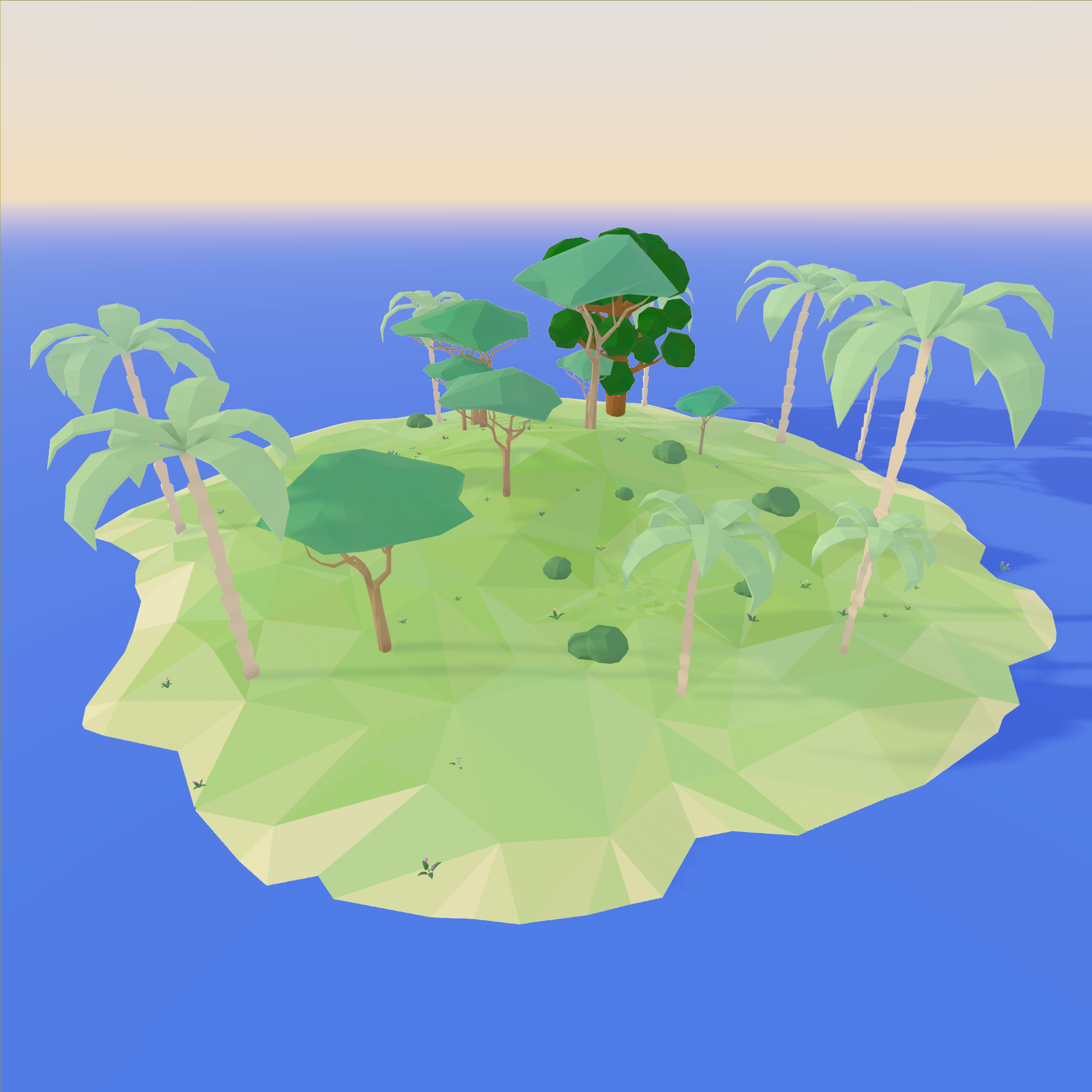}
     \end{subfigure}
     \hfill
     \begin{subfigure}[b]{0.24\textwidth}
         \centering
         \includegraphics[width=\textwidth]{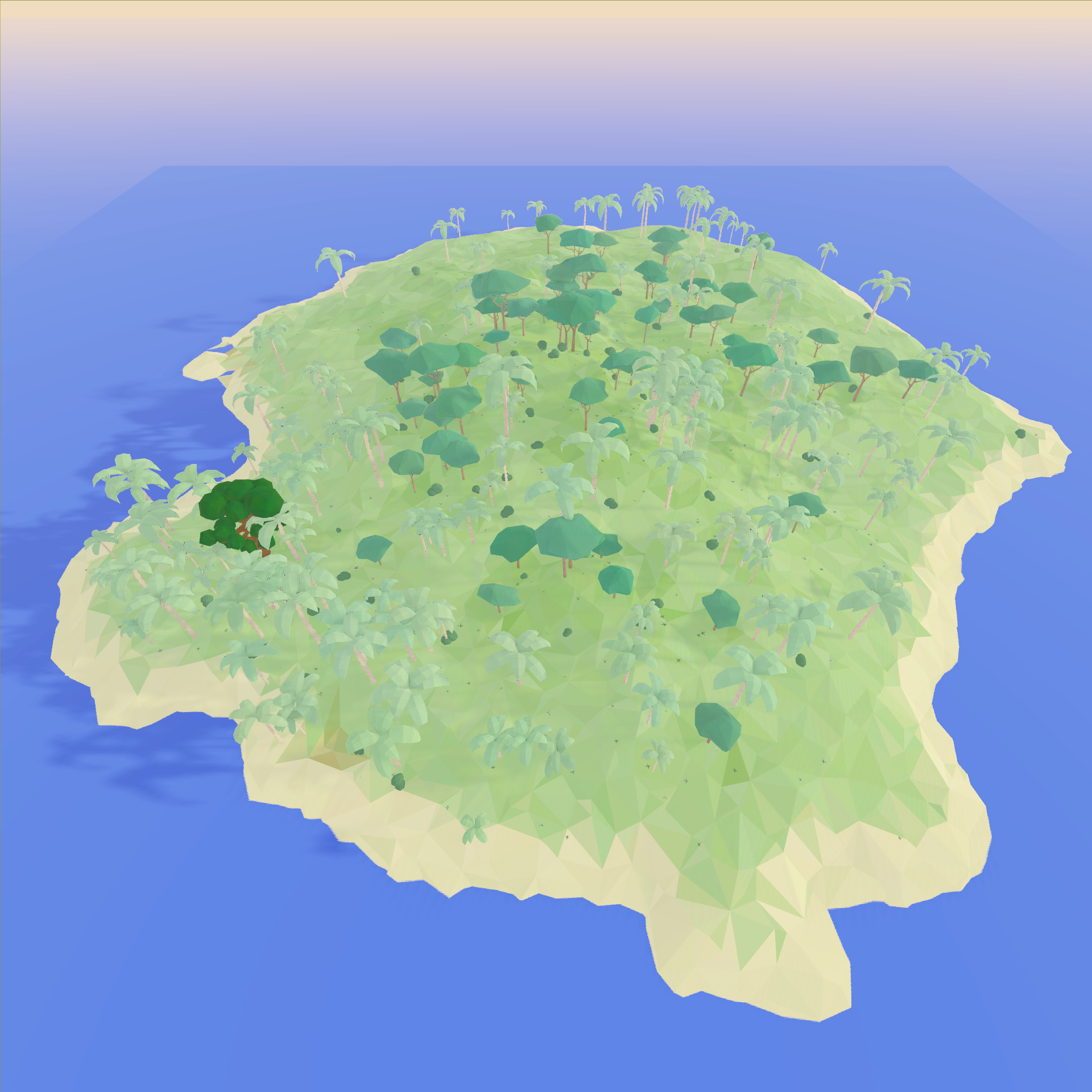}
     \end{subfigure}
     \hfill
     \begin{subfigure}[b]{0.24\textwidth}
         \centering
         \includegraphics[width=\textwidth]{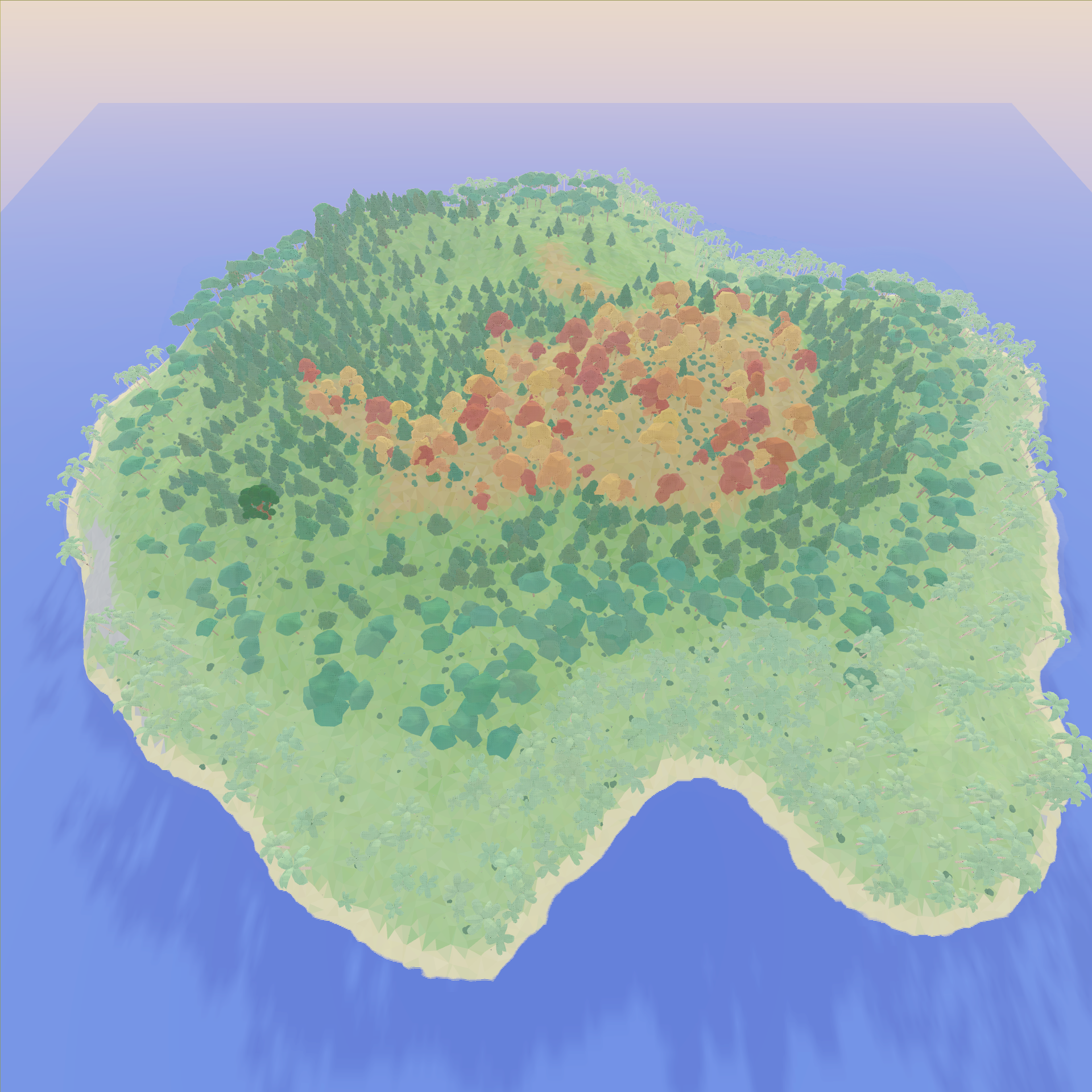}
     \end{subfigure}
        \caption{ Each of the worlds used for profiling and their sizes, from left to right: Small (32m $\times$ 32m), Medium (64m $\times$ 64m), Large (220m $\times$ 220m), and Huge (440m $\times$ 440m) }
        \label{fig:profiling_worlds}
\end{figure}

One of the major contributions of Avalon is the simulator, which was specifically designed to be high-performance. Table \ref{tab:fps_multi_process_2080} gives the performance for multiple processes on a single 2080Ti GPU, and Table \ref{tab:fps_single_process_2080} gives the performance for a single process for the same configuration. Each table gives a breakdown of performance, in terms of simulated steps per second, by both the size and complexity of the worlds being simulated, as well as by the graphical options that were enabled. The worlds range in size from Small (64m $\times$ 64m) to Huge (almost a square half-kilometer). See Figure \ref{fig:profiling_worlds} for images of the levels that were used for profiling). The graphical options include which renderer to use (GLES2 vs GLES3) and which options (fancy vs basic). For the "basic" condition, shadows were disabled, lighting was done per-vertex, and all settings were configured for the Godot defaults for mobile rendering. For the "fancy" condition, the opposite was true--lighting was per pixel, shadows were enabled, and all Godot defaults for desktop computers were used. Additionally, GLES3 has MSAA 4x enabled for both conditions, since we use this for our agent because it makes it easier to detect small objects from farther away without incurring much of a performance penalty.

Our performance numbers are meant to get as close as possible to the performance numbers reported for Habitat. We used a 2080Ti GPU in order to make our numbers more directly comparable. We also show results on a 3090 GPU (which was used for all of our other experiments) in Tables \ref{tab:fps_multi_process_3090} and \ref{tab:fps_single_process_3090}. The 2080Ti was paired with a Intel i9-10980XE CPU, while the 3090 had an AMD Ryzen 9 5950X. We also rendered at 128 x 128, a higher resolution than used in the rest of our paper, in order to be more directly comparable. Like Habitat, our benchmarking setting consisted of reading actions from a pipe and writing outputs to a file (no networks were being trained or run on the same GPU). Unlike Habitat, our simulator works in a straightforward loop of reading actions, stepping physics and gameplay logic, and rendering the resulting frame, while their fastest numbers came from an interleaved setting that delayed observation of the effect of an action by an extra frame. We also do not require any frame pointer passing or other complex integrations, making it easier to integrate our simulator into any network training setup, including distributed settings.

It should be noted that this performance is close to the maximum possible, given the hardware and software available today--a non-trivial amount of the time spent in rendering our Small world is spent simply clearing the screen via the glClear call (a necessary component for rendering). The single-process numbers represent the maximum possible speed-up over real-time for a single agent in our simulator (e.g., since we run the simulation at 10 steps per simulated second and a single process on the 3090 can do 4,114 steps per second, we can run an agent at up to 411.4x real time).  It should be noted that this speed is not reflective of performance real world--practical training and evaluation is dominated by the network forward and backward passes. The multi-process numbers are more reflective of training throughput and efficiency per-GPU.

\begin{table}
    \centering
\begin{tabular}{lrrrr}
\toprule
                    & \textbf{Small}    & \textbf{Medium}    & \textbf{Large}    & \textbf{Huge}     \\
\midrule

GLES2 (basic) &   10,157 &    9,854 &    5,886 &    1,218 \\
GLES2 (fancy) &    8,452 &    7,811 &    3,545 &      562 \\
GLES3 (basic) &    6,853 &    6,706 &    5,321 &    1,757 \\
GLES3 (fancy) &    5,901 &    5,685 &    3,627 &      848 \\

\bottomrule
\end{tabular}
    \caption{
    Multi-process performance (steps per second) on a single 2080TI GPU on various sizes of worlds (from Small to Huge).
    }
    \label{tab:fps_multi_process_2080}
\end{table}

\begin{table}
    \centering
\begin{tabular}{lrrrr}
\toprule
                    & \textbf{Small}    & \textbf{Medium}    & \textbf{Large}    & \textbf{Huge}     \\
\midrule

GLES2 (basic) &    2,880 &    2,747 &    2,144 &      623 \\
GLES2 (fancy) &    2,574 &    2,409 &    1,583 &      415 \\
GLES3 (basic) &    2,557 &    2,385 &    2,066 &      735 \\
GLES3 (fancy) &    2,203 &    2,033 &    1,506 &      519 \\

\bottomrule
\end{tabular}
    \caption{
    Single-process performance (steps per second) on a single 2080TI GPU on various sizes of worlds (from Small to Huge).
    }
    \label{tab:fps_single_process_2080}
\end{table}

\begin{table}
    \centering
\begin{tabular}{lrrrr}
\toprule
                    & \textbf{Small}    & \textbf{Medium}    & \textbf{Large}    & \textbf{Huge}     \\
\midrule

GLES2 (basic) &   11,730 &   10,963 &    5,999 &    1,120 \\
GLES2 (fancy) &    9,445 &    8,614 &    3,611 &      529 \\
GLES3 (basic) &    7,393 &    7,290 &    6,025 &    2,175 \\
GLES3 (fancy) &    6,336 &    6,182 &    4,056 &    1,081 \\

\bottomrule
\end{tabular}
    \caption{
    Multi-process performance (steps per second) on a single 3090 GPU on various sizes of worlds (from Small to Huge).
    }
    \label{tab:fps_multi_process_3090}
\end{table}

\begin{table}
    \centering
\begin{tabular}{lrrrr}
\toprule
                    & \textbf{Small}    & \textbf{Medium}    & \textbf{Large}    & \textbf{Huge}     \\
\midrule

GLES2 (basic) &    4,114 &    3,953 &    3,349 &    1,198 \\
GLES2 (fancy) &    3,624 &    3,511 &    2,450 &      570 \\
GLES3 (basic) &    3,679 &    3,529 &    3,181 &    1,381 \\
GLES3 (fancy) &    3,216 &    3,036 &    2,278 &      795 \\

\bottomrule
\end{tabular}
    \caption{
    Single-process performance (steps per second) on a single 3090 GPU on various sizes of worlds (from Small to Huge).
    }
    \label{tab:fps_single_process_3090}
\end{table}

\end{document}